\documentclass[letterpaper]{article} 
\usepackage{aaai23}  
\usepackage{times}  
\usepackage{helvet}  
\usepackage{courier}  
\usepackage[hyphens]{url}  
\usepackage{graphicx} 
\usepackage{natbib}  
\usepackage{caption} 
\usepackage{algorithm}
\usepackage{algorithmic}
\usepackage{subfig}
\usepackage{comment}
\usepackage{multirow}
\usepackage{booktabs}
\usepackage{xcolor}
\usepackage{tabularx}
\usepackage[super]{nth}
\newcolumntype{Y}{>{\centering\arraybackslash}X}
\usepackage{enumitem}
\usepackage{color,soul}
\usepackage{bbding}
\soulregister\cite7
\soulregister\ref7
\soulregister\pageref7
\usepackage{amsmath,amssymb}
\usepackage{newfloat}
\usepackage{listings}
\DeclareCaptionStyle{ruled}{labelfont=normalfont,labelsep=colon,strut=off} 
\floatstyle{ruled}
\newfloat{listing}{tb}{lst}{}
\floatname{listing}{Listing}
\title{Estimating Reflectance Layer from a Single Image: Integrating Reflectance Guidance and Shadow/Specular Aware Learning}
\author{
	Yeying Jin\textsuperscript{\rm 1},
	Ruoteng Li\textsuperscript{\rm 1,2},
	Wenhan Yang\textsuperscript{\rm 3},
	Robby T. Tan\textsuperscript{\rm 1,4}
}
\affiliations{
	\textsuperscript{\rm 1} National University of Singapore, Singapore\\
	\textsuperscript{\rm 2} ByteDance, Singapore\\
	\textsuperscript{\rm 3} Peng Cheng Laboratory, China\\
	\textsuperscript{\rm 4} Yale-NUS College, Singapore\\
	jinyeying@u.nus.edu, liruoteng@u.nus.edu, yangwh@pcl.ac.cn, robby.tan@nus.edu.sg\\
}

\begin{document}
\maketitle

\begin{abstract}
Estimating the reflectance layer from a single image is a challenging task. It becomes more challenging when the input image contains shadows or specular highlights, which often render an inaccurate estimate of
the reflectance layer. Therefore, we propose a two-stage learning method, including reflectance guidance and a Shadow/Specular-Aware (S-Aware) network to tackle the problem. In the first stage, an initial
reflectance layer free from shadows and specularities is obtained with the constraint of novel losses that are guided by prior-based shadow-free and specular-free images. To further enforce the reflectance
layer to be independent of shadows and specularities in the second-stage refinement, we introduce an S-Aware network that distinguishes the reflectance image from the input image. Our network employs a
classifier to categorize shadow/shadow-free, specular/specular-free classes, enabling the activation features to function as attention maps that focus on shadow/specular regions. Our quantitative and
qualitative evaluations show that our method outperforms the state-of-the-art methods in the reflectance layer estimation that is free from shadows and specularities. Code is at: \url{https://github.com/jinyeying/S-Aware-network}.
\end{abstract}

\section{Introduction}
\label{sec:intr}
Reflectance layer estimation is a fundamental task in computer vision. 
It is a part of intrinsic image decomposition, which decomposes an input image into the reflectance layer and the shading layer~\cite{barrow1978recovering}. 
The reflectance layer provides useful information for real-world applications, \textit{e.g.}, surface retexturing~\cite{garces2012intrinsic}, relighting~\cite{liu2020learning}, object compositing~\cite{bi20151}.
Unfortunately, reflectance layer estimation from a single image is an inherently ill-posed problem.

\begin{figure}[t]
	\centering
	\captionsetup[subfloat]{labelformat=empty}
	\captionsetup[subfloat]{farskip=1pt}
	\subfloat[Input]{\includegraphics[width=0.245\columnwidth]{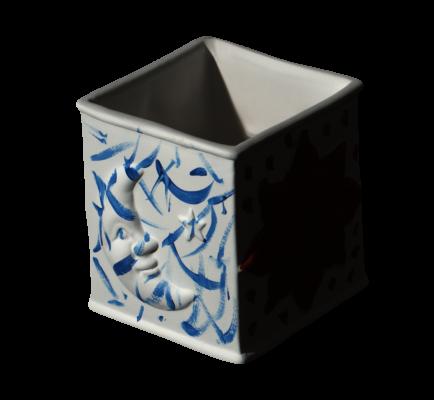}}\hfill
	\subfloat[Ours]{\includegraphics[width=0.245\columnwidth]{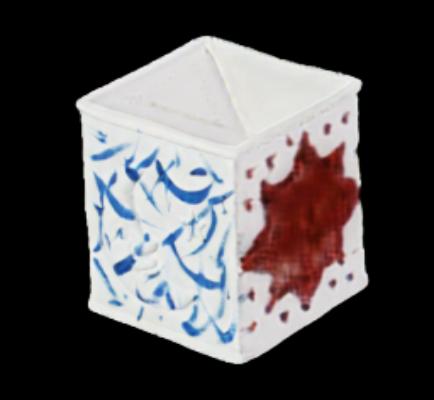}}\hfill
	\subfloat[Ground Truth]{\includegraphics[width=0.245\columnwidth]{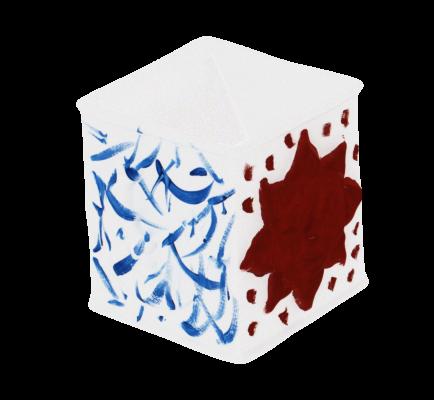}}\hfill
	\subfloat[PIENet~\shortcite{das2022pie}]{\includegraphics[width=0.245\columnwidth]{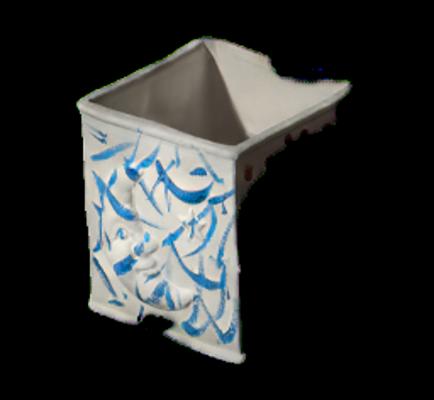}}\hfill
	\subfloat[Input]{\includegraphics[width=0.245\columnwidth]{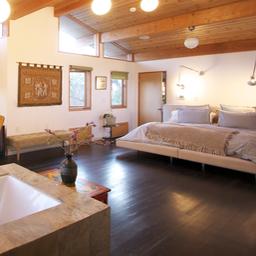}}\hfill
	\subfloat[Ours]{\includegraphics[width=0.245\columnwidth]{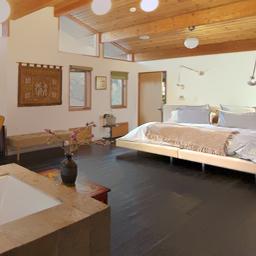}}\hfill
	\subfloat[UID~\shortcite{zhang2021unsupervised}]{\includegraphics[width=0.245\columnwidth]{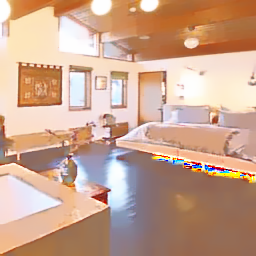}}\hfill
	\subfloat[PIENet~\shortcite{das2022pie}]{\includegraphics[width=0.245\columnwidth]{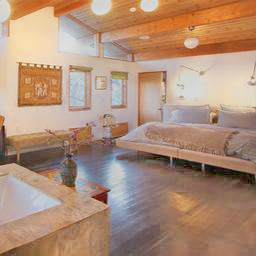}}\hfill
	\caption{Comparison between our results and the state-of-the-art methods UIDNet~\cite{zhang2021unsupervised} and PIENet~\cite{das2022pie}. Unlike existing methods, our method estimates the reflectance layer free from shadows and specularities.}
	\label{fig:intro}
\end{figure}

Several methods have been proposed to deal with the problem. 
Non-deep learning methods \textit{e.g.},~\cite{bell2014intrinsic} impose priors on the estimated reflectance layer. 
Supervised learning methods \textit{e.g.},~\cite{das2022pie}, while effective, still suffer from problems, particularly those related to shadows and specularities.
Some existing unsupervised learning methods learn from time-lapse data~\cite{li2018learning,liu2020learning}, multi-view data~\cite{yi2020leveraging} or sets of unpaired images~\cite{liu2020unsupervised}.

Among these methods, only few consider shadows~\cite{baslamisli2021shadingnet,baslamisli2021physics} and specular highlights~\cite{yi2020leveraging}.
However, they suffer from a few drawbacks.
First, most of them are trained only on synthesized datasets, and thus  their accuracy depends heavily on the quality of the synthesized data.
Second, while these methods work well in some cases, they tend to fail in handling large shadow or specular regions, as shown in Fig.~\ref{fig:intro}.
Third, most of these methods assume that reflectance changes cause abrupt image intensity changes, while shading changes cause smooth image intensity changes~\cite{land1971lightness}.
However, shadows, which should belong to the shading layer, can also cause abrupt intensity changes in the real world, which goes against this assumption.

In this paper, our goal is to decompose a diffuse reflectance layer that is free from shadows and specularities.
To achieve this, we introduce a two-stage network based on reflectance guidance and a Shadow/Specular-Aware (S-Aware) network.
The first stage is to obtain an initial diffuse reflectance layer free from shadows and specularities.
To achieve this, we introduce novel losses constrained by prior-based shadow-free~\cite{Finlayson04} and specular-free images~\cite{tan2005specular}.

Once the initial reflectance layer is obtained, we feed it into the S-Aware network, which identifies shadow/specular regions by modulating the activation features as attention maps.
To enable the S-Aware network to self-learn the shadow/specular regions, a Shadow/Specularity Classifier (S-Classifier) is employed to distinguish the first-stage estimated reflectance layer from the input image.
Specifically, by distinguishing shadows from shadow-free or specularities from specular-free, the activation features form shadow/specular attention.
Our attention mechanism modulates the activation weights with encoded features that capture spatially varying regions.
Once our network can focus on shadow/specular regions, it can refine the reflectance layer to be free from shadows and specularities.
In summary, here are our contributions:
\begin{itemize}[noitemsep,topsep=1pt]
	\item To the best of our knowledge, our method is the first single-image diffuse reflectance layer estimation network that performs robustly even in the presence of shadows and specularities. This differs from existing intrinsic image decomposition methods, which tend to fail to remove shadows and specularities from the reflectance layer.
	\item We introduce a reflectance guidance framework that provides reliable guidance for our network to learn the reflectance layer. Our novel losses constrain the reflectance layer free from shadows and specularities, based on prior-based shadow-free and specular-free images.
	\item We propose an S-Aware network to modulate activation features as shadow/specular attention. 
	Our network automatically learns to focus on these regions, since our S-Classifier distinguishes shadows from shadow-free or specularities from specular-free.
\end{itemize}
Our quantitative and qualitative evaluations show that our method outperforms the state-of-the-art methods in various datasets 
for suppressing shadows and specularities in the reflectance layer, including: 4 intrinsic datasets, 2 shadow datasets and 1 specular dataset.

\begin{figure*}[t]
	\centering
	\captionsetup[subfloat]{farskip=1pt}
	\subfloat{\includegraphics[width=1\textwidth]{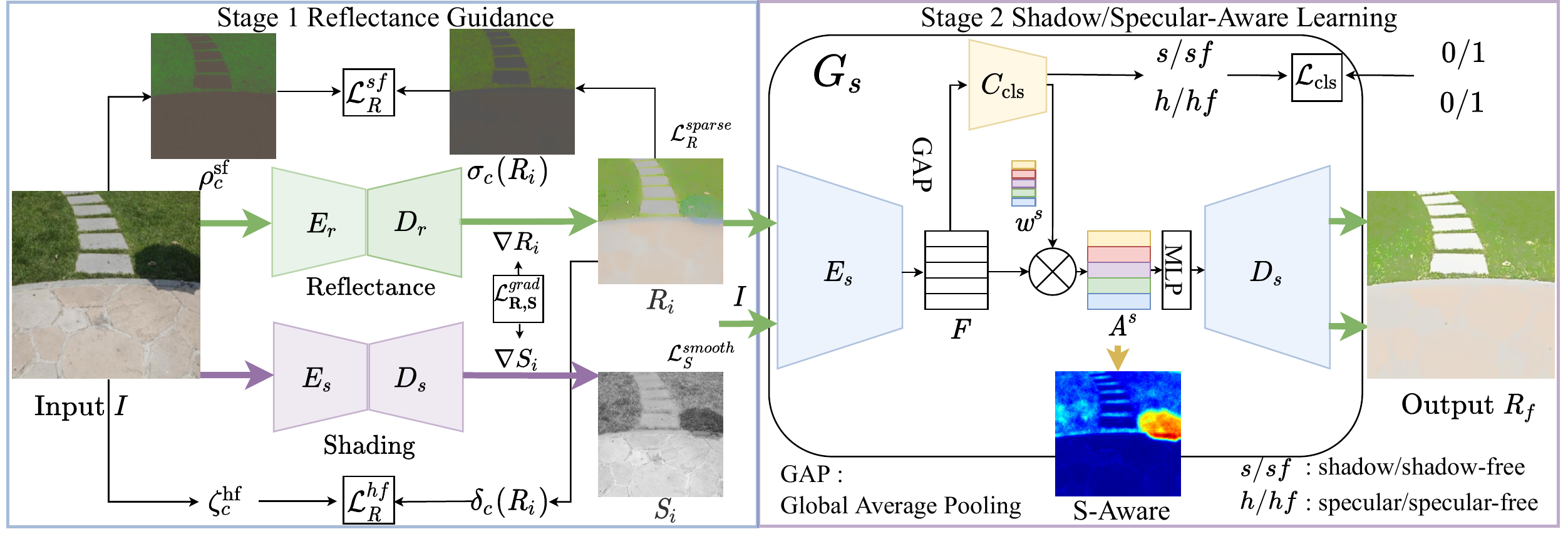}}
	\caption{Our framework consists of two stages: reflectance guidance and Shadow/Specular-Aware (S-Aware).
		In the first stage, to obtain the initial reflectance layer $\mathit{R}_i$, we propose the novel shadow-free $\mathcal{L}_\mathit{R}^\text{sf}$ (see Fig.~\ref{fig:chromaticityloss}) and specular-free $\mathcal{L}_\mathit{R}^\text{hf}$ losses (see Fig.~\ref{fig:sploss}).
		In the second stage, the initial reflectance layer and the input image are fed to the S-Aware network to obtain the final reflectance output $\mathit{R}_f$.
		Our S-Aware network is represented by $G_s$ (contain an S-Classifier $C_\text{cls}$).
		By distinguishing shadows from shadow-free, or specularities from specular-free, our network automatically learns the activation weights $\mathit{w}^s$. Multiplying $\mathit{w}^s$ to modulate the encoded features $\mathit{F}$, obtain S-Aware attention $\mathit{A}^s$, which focuses on shadow/specular regions.}
	\label{fig:model}
\end{figure*}

\section{Related Work}
\label{sec:related_work}
\noindent \textbf{Reflectance Layer Estimation}
The Retinex algorithm~\cite{land1971lightness} assumes reflectance changes cause large gradients, while shading variation corresponds to small gradients.
Subsequently, various priors, \textit{e.g.}, sparse reflectance, reflectance colors, textures, depth are utilized to regularize the reflectance layer.
However, with only hand-crafted constraints, these methods are not adaptive enough, and scene-specific parameters are hard to cover real-world shadows and specularities.
Moreover, the assumption that shading changes cause smooth image intensity changes is not likely to cover the large shading change, such as shadow regions.

In recent years, many deep-learning-based methods have been introduced~\cite{garces2022survey}, and most of them adopt supervised
learning~\cite{narihira2015direct,baslamisli2018cnn,fan2018revisiting,das2022pie}.
The challenge of applying learning-based methods is the lack of a variety of real images with ground truth. 
Synthetic datasets \textit{e.g.},~\cite{butler2012naturalistic,shi2017learning,li2018cgintrinsics,baslamisli2021shadingnet,han2022blind} highly depend on the quality of the rendering techniques and 3D models. 
Poor rendering quality will make the network fail to handle real images in the testing stage, due to the gaps between the synthetic and real image domains.

Real image datasets \textit{e.g.},~\cite{grosse2009ground,bell2014intrinsic,kovacs2017shading}) are either too limited, lack diversity, or have highly sparse annotations.
To address the limitations of supervised learning methods, a few unsupervised learning methods are proposed.
They mainly focus on images with static reflectance and varied illumination~\cite{lettry2018unsupervised,ma2018single,li2018learning,liu2020learning}.
Unlike multi-image approaches, unpaired translation~\cite{liu2020unsupervised} and internal self-similarity~\cite{zhang2021unsupervised} provide more solutions.
However, most existing methods have artefacts in the reflectance layer, particularly on the shadow and specular regions. 
Unlike existing methods, our network estimates the reflectance layer that is free from shadows and specularities.

\vspace{0.2cm}
\noindent \textbf{Shadow and Specularity Removal}
A few intrinsic image decomposition methods~\cite{baslamisli2021physics,baslamisli2021shadingnet,zhang2021unsupervised} attempt to address shadows, and other methods~\cite{shi2017learning,yi2020leveraging,li2020inverse} attempt to solve specular highlights separation using a large-scale synthetic data in training. 
However, the accuracy of the rendering limits the performance of the methods on the real-world shadows and specularities.

Note that, there are a few shadow/specularity removal methods that do not decompose the input image to the intrinsic layers, such as
~\cite{kawakami2005consistent,jin2021dc,guo2023shadowformer,jin2022shadowdiffusion}) for shadow removal, 
and
~\cite{tan2003illumination,tan2005specular,yang2010real,shen2013real,guo2018single} for specularity removal.
While these methods work to some extent, the problems of shadow/specular removal are still open problems.

\section{Proposed Method}
\label{sec:method}
Fig.~\ref{fig:model} shows our pipeline to estimate the reflectance layer that is free from shadows/specularities. First, we obtain the initial reflectance layer by employing reflectance guidance, and then refine the reflectance layer using the Shadow/Specular-Aware (S-Aware) network.

\begin{figure}[t]
	\centering
	\captionsetup[subfloat]{labelformat=empty}
	\subfloat{\includegraphics[width=0.47\textwidth]{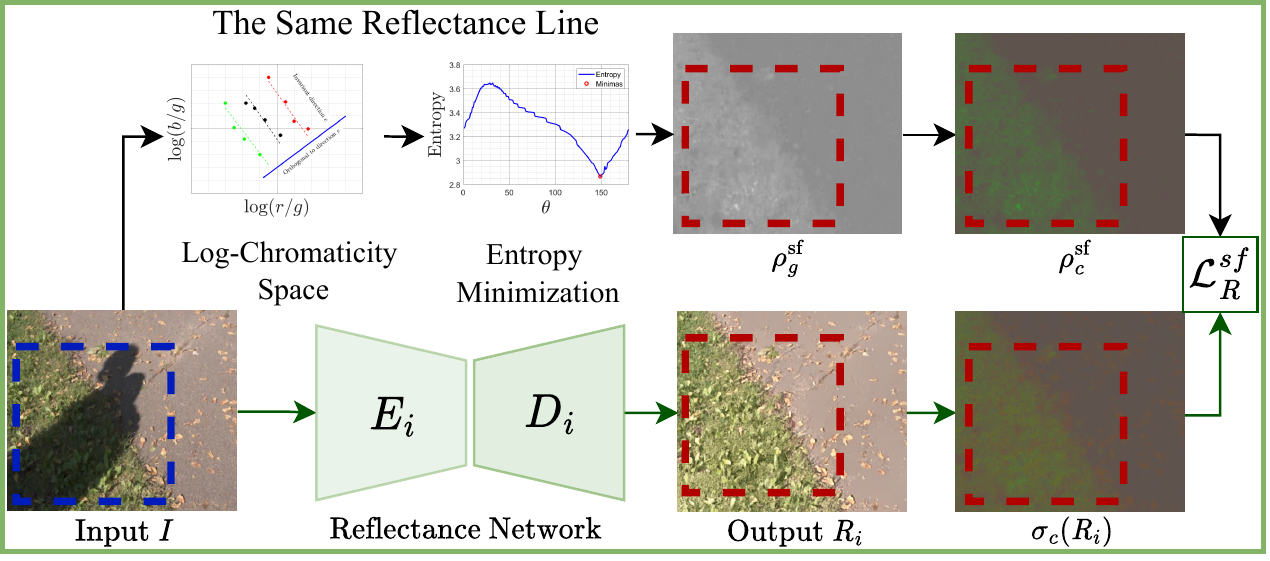}}
	\caption{Using the shadow-free loss $\mathcal{L}_\mathit{R}^\text{sf}$, our first stage reflectance network learns to remove shadows in the reflectance layer $\mathit{R}_i$. Shadow-free image $\rho_c^\text{sf}$ do not have shadow regions (see the red dashed rectangles).}
	\label{fig:chromaticityloss}
\end{figure}

\begin{figure}[t]
	\centering
	\captionsetup[subfloat]{labelformat=empty}
	\subfloat{\includegraphics[width=0.47\textwidth]{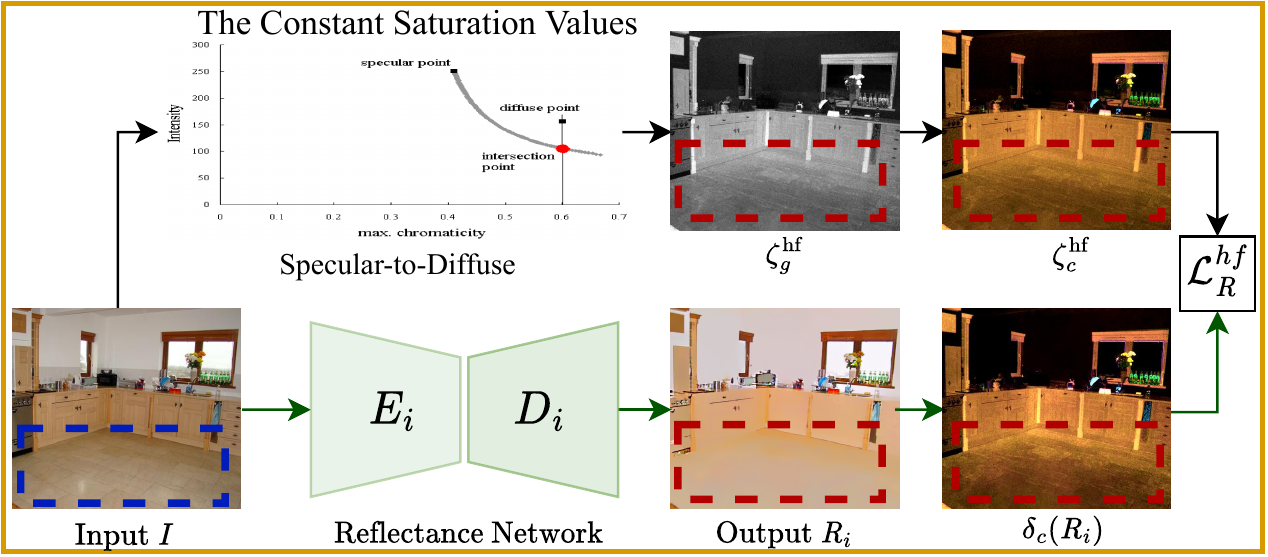}}
	\caption{Using the specular-free loss $\mathcal{L}_\mathit{R}^\text{hf}$, our first stage reflectance network learns to remove specularities in the reflectance layer $\mathit{R}_i$. Specular-free image $\zeta_c^\text{hf}$ is free from specularities (see the red dashed rectangles).}
	\label{fig:sploss}
\end{figure}

\subsection{Intrinsic Image Reflectance Guidance}
We estimate the reflectance layer $\mathit{R}$, given an input image $\mathit{I}$ and governed by $\mathit{I} = \mathit{R} \odot \mathit{S}$, where $\odot$ is the element-wise multiplication. $\mathit{S}$ is the shading layer.
Note that, shadows naturally belong to the shading layer, as they are phenomena caused by light (instead of by objects). 
While specularities are caused by surface reflectance~\cite{tan2021specularity}, and thus should belong to the reflectance layer.
However, many applications assume diffuse only reflectance, and hence the reflectance layer that is independent from specularities is more desirable.
Therefore, given an input image, our goal is to obtain the diffuse reflectance layer that is free from both shadows and specular highlights.
Throughout this paper, reflectance layer refers to the diffuse reflectance layer. 
We employ encoder-decoder ($E_r$-$D_r$ and $E_s$-$D_s$, with subscript $r$, $s$ stands for reflectance, shading) to simultaneously decompose the input image $\mathit{I}$ into an initial reflectance layer $\mathit{R}_i$ and an initial shading layer $\mathit{S}_i$, as shown in Fig.~\ref{fig:model}.

Due to the ill-posed nature of our problem, we propose to utilize the shadow-free image~\cite{Finlayson04} and the specular-free image~\cite{tan2005specular} as priors to guide the first stage decomposition network.
For this guidance, we propose two novel losses: shadow-free loss $\mathcal{L}_\mathit{R}^\text{sf}$ and specular-free loss $\mathcal{L}_\mathit{R}^\text{hf}$.

\begin{figure}[t!]
	\captionsetup[subfloat]{farskip=1pt}
	\centering
	\subfloat{\includegraphics[width = 0.243\columnwidth]{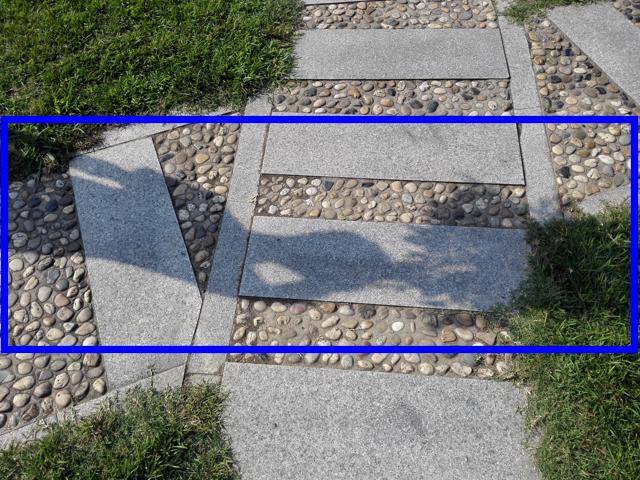}}\hfill
	\subfloat{\includegraphics[width = 0.243\columnwidth]{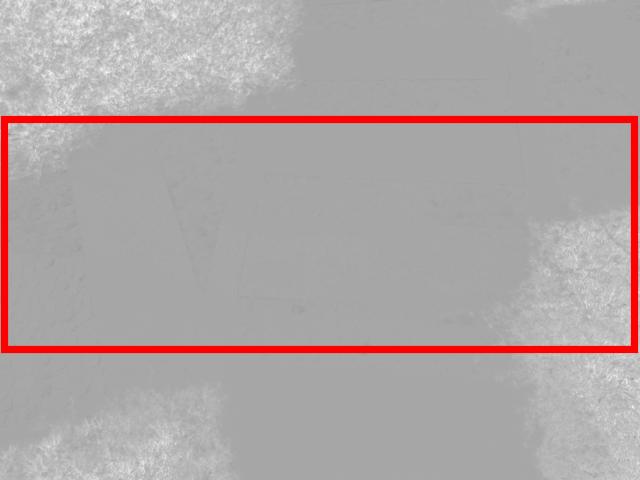}}\hfill
	\subfloat{\includegraphics[width = 0.243\columnwidth]{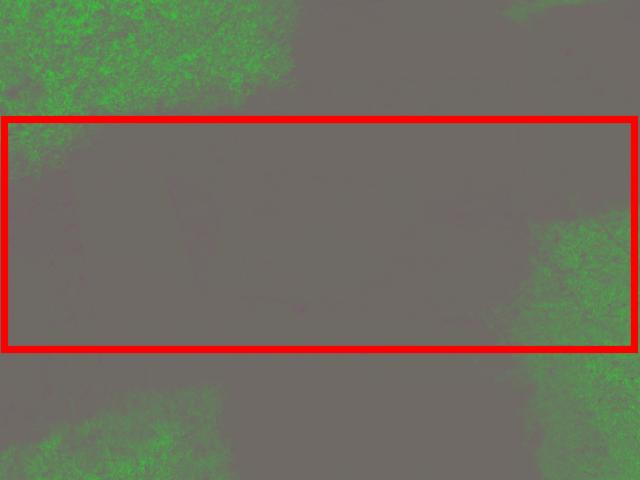}}\hfill
	\subfloat{\includegraphics[width = 0.243\columnwidth]{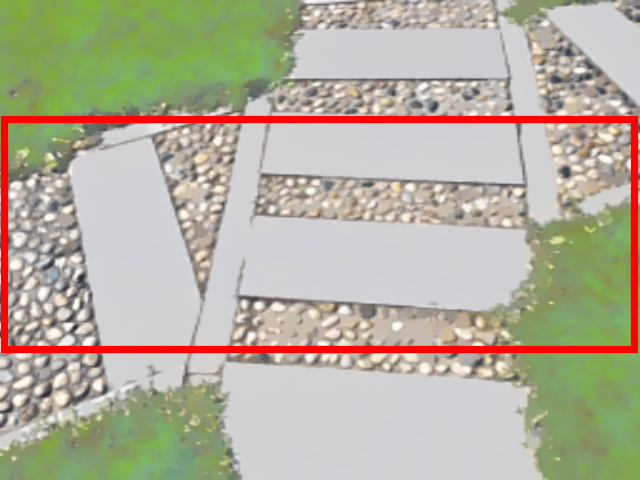}}\hfill\\
	\setcounter{subfigure}{0}
	\subfloat[Input $\mathit{I}$]{\includegraphics[width = 0.243\columnwidth]{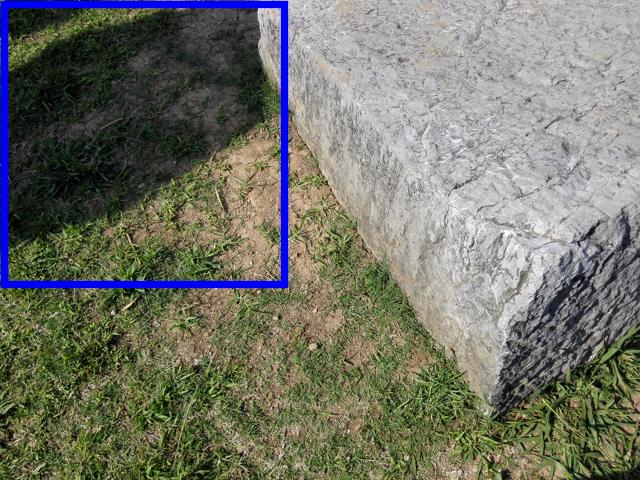}}\hfill
	\subfloat[$\rho_g^\text{sf}$\label{fig:sfg}]{\includegraphics[width = 0.243\columnwidth]{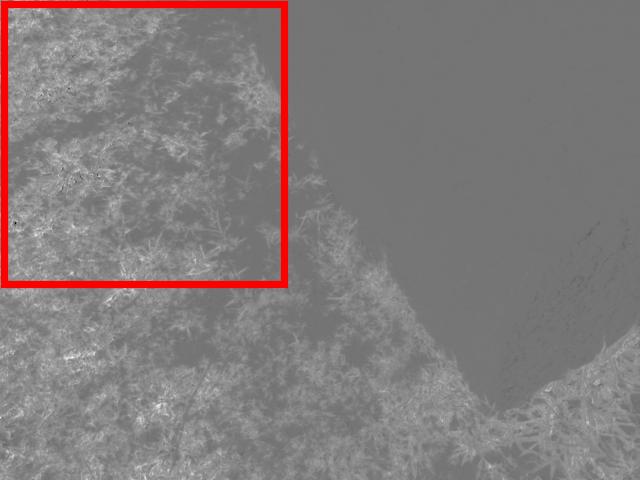}}\hfill
	\subfloat[$\rho_c^\text{sf}$\label{fig:sfc}]{\includegraphics[width = 0.243\columnwidth]{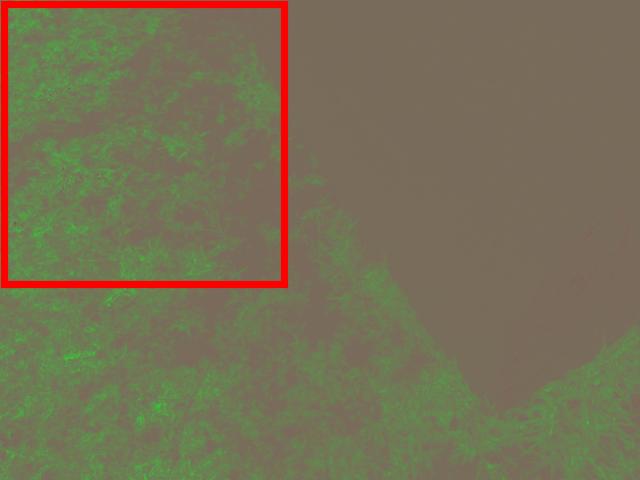}}\hfill
	\subfloat[$\mathit{R}_i$]{\includegraphics[width = 0.243\columnwidth]{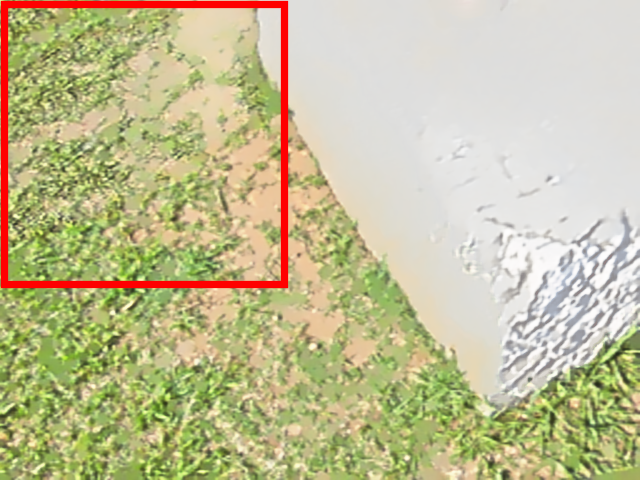}}\hfill\\
	\caption{(a) Input image with shadows. (b) The grayscale shadow-free image, and (c) The colored shadow-free image $\rho_c^\text{sf}$ do not have shadows (see the red boxes), providing prior guidance to our network. Therefore, our reflectance network learns to obtain (d) the shadow-free reflectance layer $\mathit{R}_i$.}
	\label{fig:chromaticity}
\end{figure}

\vspace{0.3cm}
\noindent \textbf{Shadow-Free Loss}
To obtain a shadow-free reflectance layer, we compute a shadow-free image in the log-chromaticity space~\cite{Finlayson09}. 
In this space, pixels belonging to the same reflectance surface form a single line, where the line is dependent on light colors (implying that shadow and non-shadow pixels of the same reflectance will lie on the same line).
Entropy minimization can capture this line and generate a grayscale image free from shadows (Fig.~\ref{fig:sfg}).
Then, we compute a colored shadow-free image, $\rho_c^\text{sf}$, by instilling the illumination colors back~\cite{drew2003recovery} (Fig.~\ref{fig:sfc}).

Guided by the colored shadow-free image, $\rho_c^\text{sf}$, our reflectance network learns to obtain the shadow-free reflectance layer through the shadow-free loss, defined as:
\begin{align}
	\mathcal{L}_\mathit{R}^\text{sf} = |\sigma_c(\mathit{R}_i) - \rho_c^\text{sf}|_{1},
	\label{eq:sf_loss}
\end{align}
where $\sigma_c$ is the chromaticity~\cite{jin2022structure}, $\sigma_c(\mathit{x}) = \frac{I_c(\mathit{x})}{I_r(\mathit{x}) + I_g(\mathit{x}) + I_b(\mathit{x})}$, $c \in\{r,g,b\}$ represents each RGB color channel.

\vspace{0.3cm}
\noindent \textbf{Specular-Free Loss}
Tan and Ikeuchi~\cite{tan2005specular} define a specular-free image as an image that has an identical geometric profile to the diffuse component of the input image yet has different hue values.
The key idea in obtaining the specular-free image is to force the saturation values to be constant for all pixels regardless of whether they are affected by specularities or not~\cite{tan2005specular,li2018robust}. 
These constant saturation values make specular highlights disappear from the image.

\begin{figure}[t!]
	\centering
	\captionsetup[subfloat]{labelformat=empty}
	\captionsetup[subfloat]{farskip=1pt}
	\rotatebox{90}{\small \phantom{z}{(a)} Input}\hspace{0.1cm}
	\subfloat{\includegraphics[width = 0.229\columnwidth,height=1.5cm]{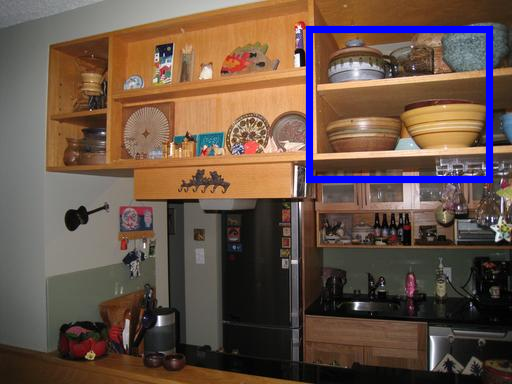}}\hfill
	\subfloat{\includegraphics[width = 0.229\columnwidth,height=1.5cm]{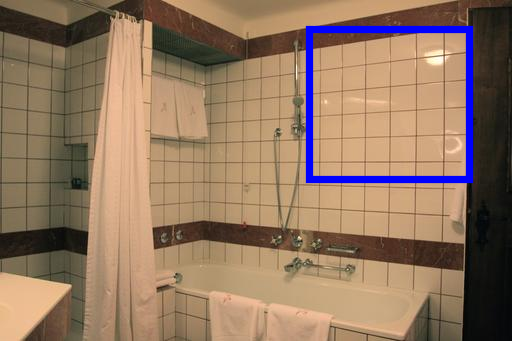}}\hfill
	\subfloat{\includegraphics[width = 0.229\columnwidth,height=1.5cm]{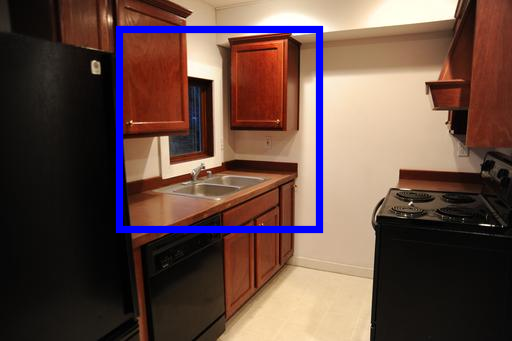}}\hfill
	\subfloat{\includegraphics[width = 0.229\columnwidth,height=1.5cm]{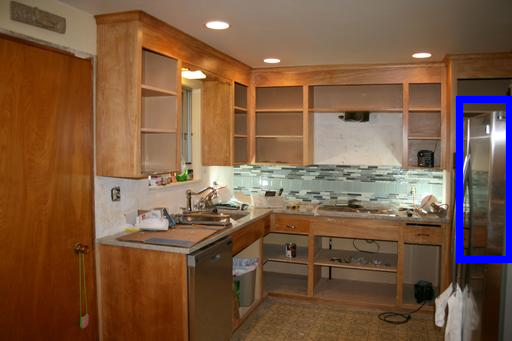}}\hfill\\	
	\rotatebox{90}{\small \phantom{zz}{(b)} $\zeta_g^\text{hf}$}\hspace{0.02cm}
	\subfloat{\includegraphics[width = 0.229\columnwidth,height=1.5cm]{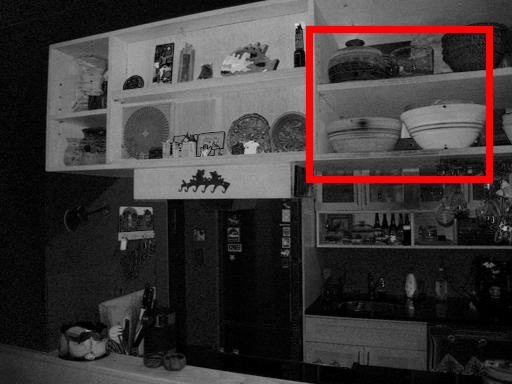}}\hfill
	\subfloat{\includegraphics[width = 0.229\columnwidth,height=1.5cm]{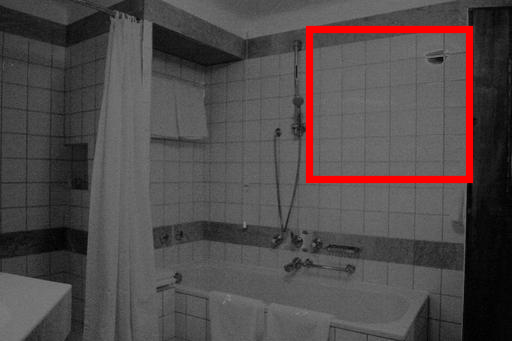}}\hfill
	\subfloat{\includegraphics[width = 0.229\columnwidth,height=1.5cm]{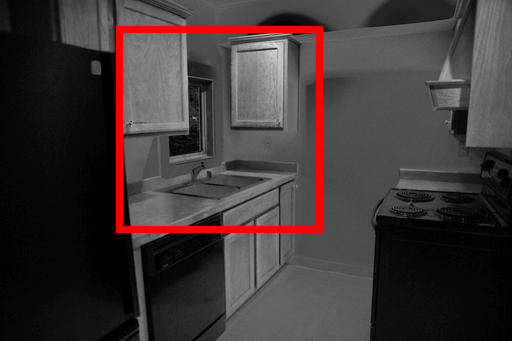}}\hfill
	\subfloat{\includegraphics[width = 0.229\columnwidth,height=1.5cm]{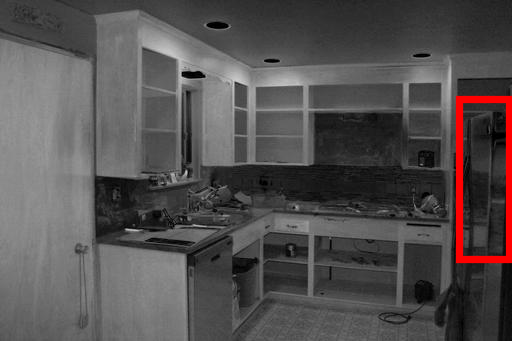}}\hfill\\
	\rotatebox{90}{\small \phantom{zz}{(c)} $\zeta_c^\text{hf}$}\hspace{0.05cm}
	\subfloat{\includegraphics[width = 0.229\columnwidth,height=1.5cm]{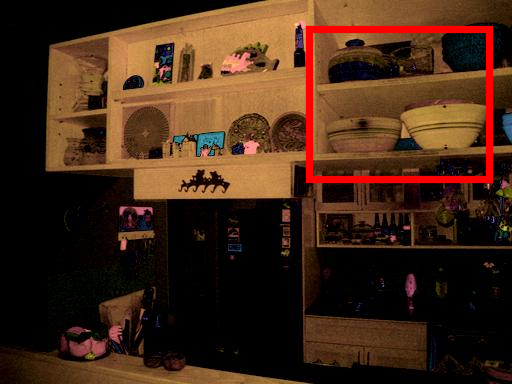}}\hfill
	\subfloat{\includegraphics[width = 0.229\columnwidth,height=1.5cm]{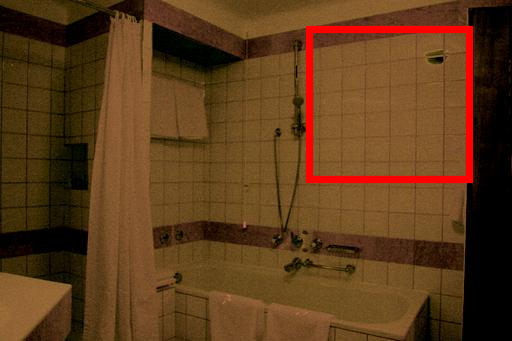}}\hfill
	\subfloat{\includegraphics[width = 0.229\columnwidth,height=1.5cm]{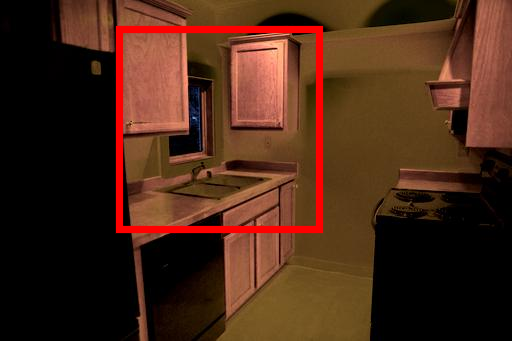}\label{fig:hf}}\hfill
	\subfloat{\includegraphics[width = 0.229\columnwidth,height=1.5cm]{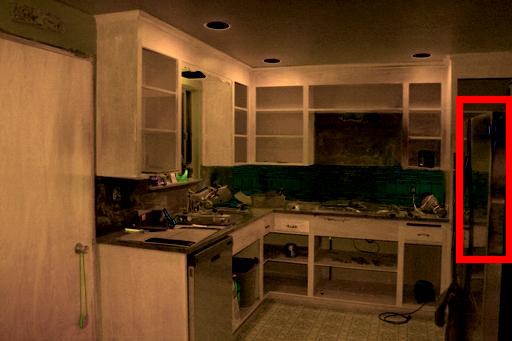}}\hfill\\
	\caption{(a) Input image with specularities, (b) The grayscale specular-free image, and (c) The colored specular-free image $\zeta_c^\text{hf}$. The specularities are reduced in the specular-free images (see the red boxes), providing prior guidance to our reflectance network.}
	\label{fig:sp}
\end{figure}

Fig.~\ref{fig:sp} shows some examples of specular-free images.
The specularities are removed in the grayscale specular-free images (second row) and the colored specular-free images $\zeta_c^\text{hf}$ (bottom row). 
Guided by the colored specular-free image, the reflectance network learns to obtain the specular-free reflectance layer through the specular-free loss, defined as:
\begin{align}
	\mathcal{L}_\mathit{R}^\text{hf} = |\delta_c(\mathit{R}_i) - \zeta_c^\text{hf}|_{1},
	\label{eq:hf_loss}
\end{align}where $\delta_c$ is the transformation from RGB to the colored specular-free images.

\vspace{0.3cm}
\noindent \textbf{Gradient Separation Loss} 
Since reflectance gradients $\nabla\mathit{R}_i$ are sparse in the gradient domain, while shading gradients $\nabla\mathit{S}_i$ are smooth~\cite{bonneel2014interactive} in the intensity domain, to separate the two uncorrelated layers, we use the gradient separation loss~\cite{zhang2018single}:
\begin{align}
	\small
	\mathcal{L}_\mathit{R,S}^\text{grad} = \sum_{n=1}^{3}\big\lVert \mathit{\tanh}(\lambda_{\mathit{R}}\lvert\nabla\mathit{R}_i^{\downarrow\!n}\rvert)\odot \mathit{\tanh}(\lambda_{\mathit{S}}\lvert\nabla\mathit{S}_i^{\downarrow\!n}\rvert)\big\rVert_{F},
	\nonumber
	\label{eq:gradientloss}
\end{align}
where $\lVert\cdot\rVert_{F}$ is the Frobenius norm, $\odot$ is the Hadamard multiplication. $\mathit{R}_i^{\downarrow\!n}$ and $\mathit{S}_i^{\downarrow\!n}$ represent $\mathit{R}_i$ and $\mathit{S}_i$  downsampled by a factor of $2^{n-1}$, $n=3$, using bilinear interpolation, the parameters $\lambda_{\mathit{R}}\!=\!\sqrt{{\lVert\nabla\mathit{S}_i^{\downarrow\!n}\rVert_F}/{\lVert\nabla\mathit{R}_i^{\downarrow\!n}\rVert_F}}$, $\lambda_{\mathit{L}}\!=\!\sqrt{{\lVert\nabla\mathit{R}_i^{\downarrow\!n}\rVert_F}/{\lVert\nabla\mathit{S}_i^{\downarrow\!n}\rVert_F}}$ are normalization factors to balance gradient magnitudes of the reflectance layer and the shading layer. 

\begin{figure}[t!]
	\captionsetup[subfloat]{labelformat=empty}
	\captionsetup[subfloat]{farskip=1pt}
	\centering
	\subfloat{\includegraphics[width = 0.162\columnwidth,height=1.2cm]{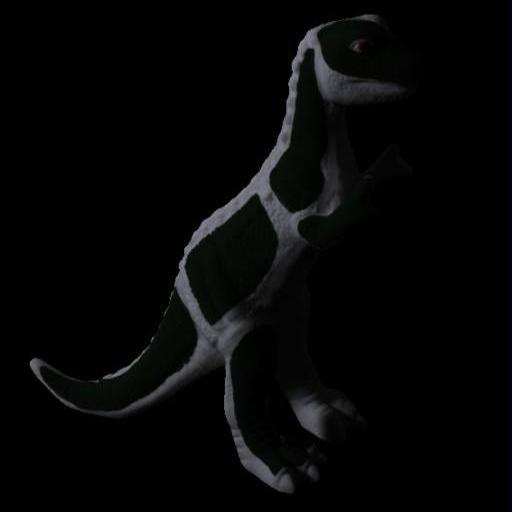}}\hfill
	\subfloat{\includegraphics[width = 0.162\columnwidth,height=1.2cm]{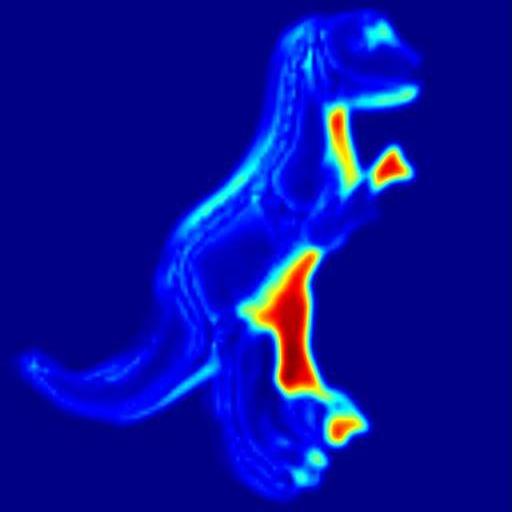}}\hfill
	\subfloat{\includegraphics[width = 0.162\columnwidth,height=1.2cm]{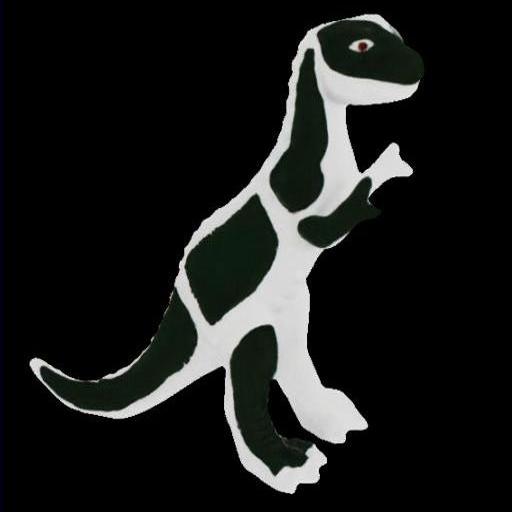}}\hfill
	\subfloat{\includegraphics[width = 0.162\columnwidth,height=1.2cm]{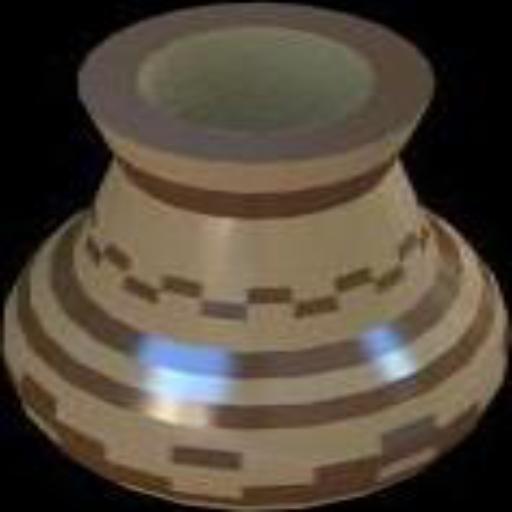}}\hfill
	\subfloat{\includegraphics[width = 0.162\columnwidth,height=1.2cm]{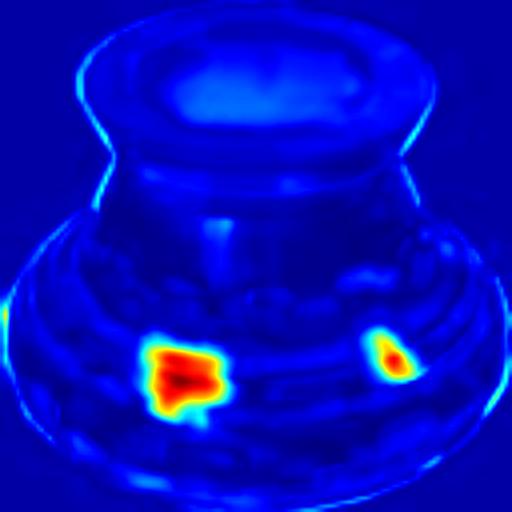}}\hfill
	\subfloat{\includegraphics[width = 0.162\columnwidth,height=1.2cm]{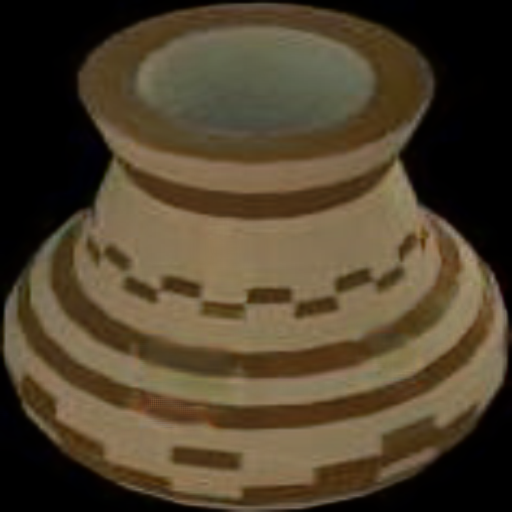}}\hfill\\
	\subfloat{\includegraphics[width = 0.162\columnwidth,height=1.2cm]{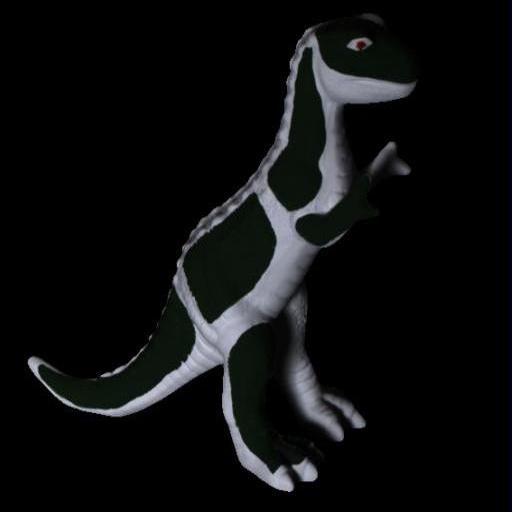}}\hfill
	\subfloat{\includegraphics[width = 0.162\columnwidth,height=1.2cm]{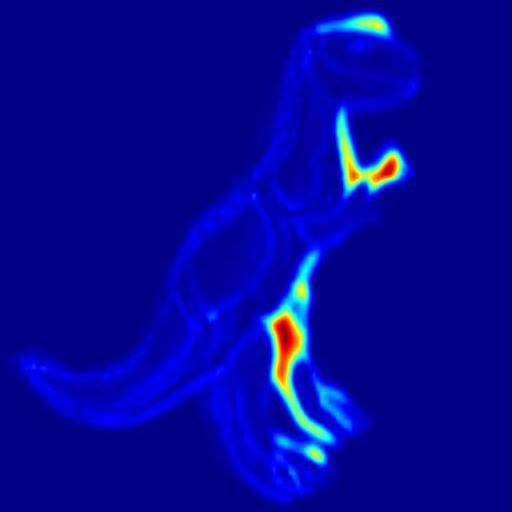}}\hfill
	\subfloat{\includegraphics[width = 0.162\columnwidth,height=1.2cm]{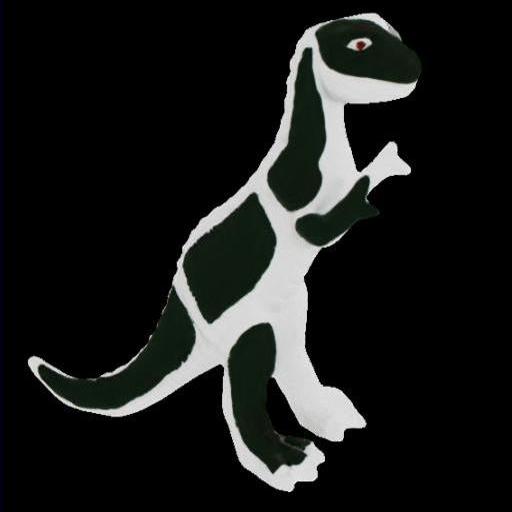}}\hfill
	\subfloat{\includegraphics[width = 0.162\columnwidth,height=1.2cm]{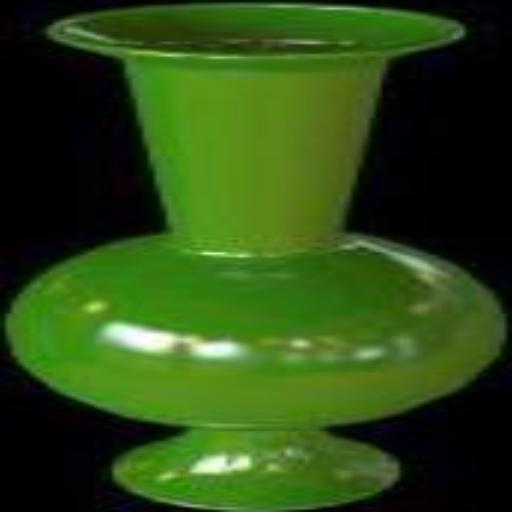}}\hfill
	\subfloat{\includegraphics[width = 0.162\columnwidth,height=1.2cm]{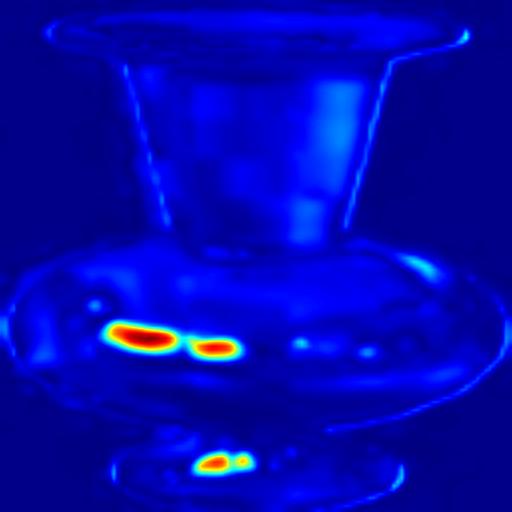}}\hfill
	\subfloat{\includegraphics[width = 0.162\columnwidth,height=1.2cm]{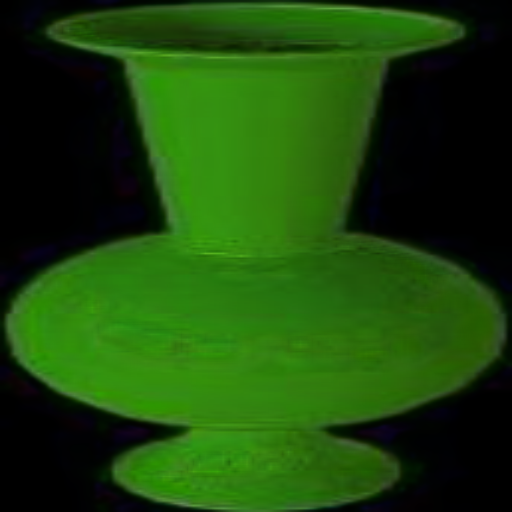}}\hfill\\	
	\subfloat{\includegraphics[width = 0.162\columnwidth,height=1.2cm]{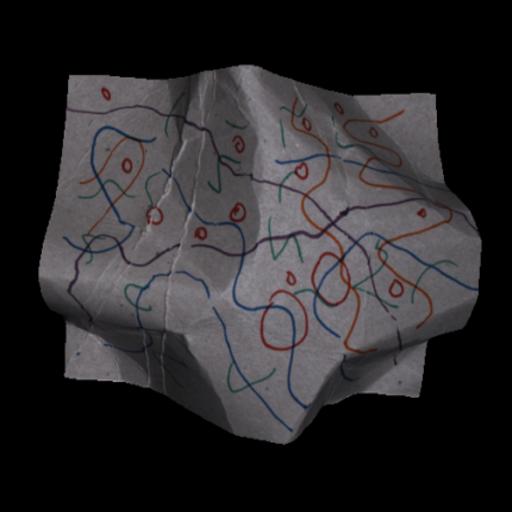}}\hfill
	\subfloat{\includegraphics[width = 0.162\columnwidth,height=1.2cm]{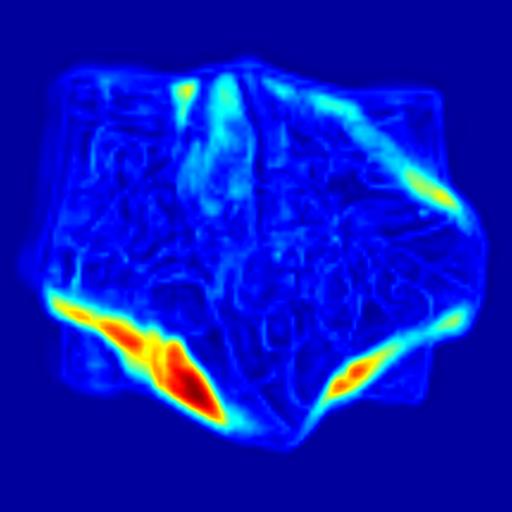}}\hfill
	\subfloat{\includegraphics[width = 0.162\columnwidth,height=1.2cm]{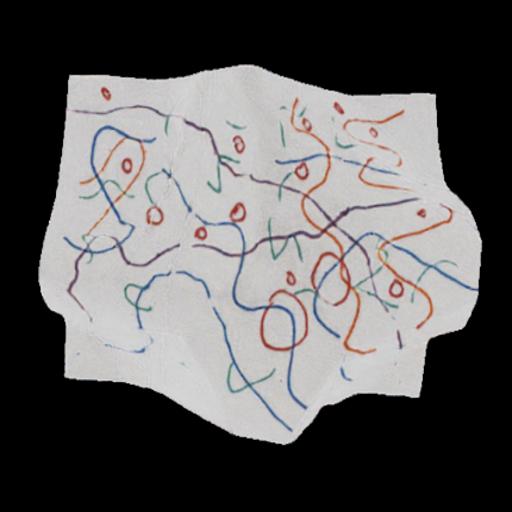}}\hfill
	\subfloat{\includegraphics[width = 0.162\columnwidth,height=1.2cm]{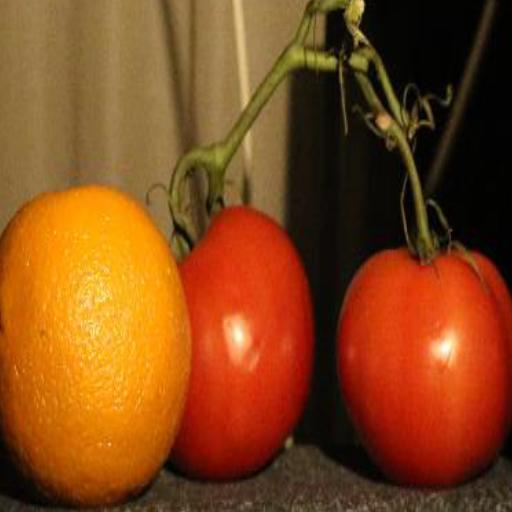}}\hfill
	\subfloat{\includegraphics[width = 0.162\columnwidth,height=1.2cm]{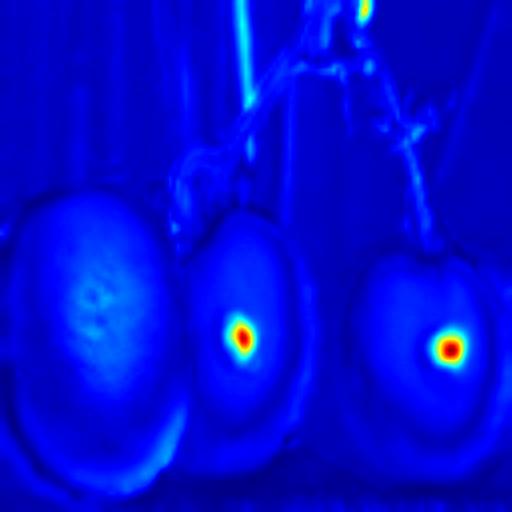}}\hfill
	\subfloat{\includegraphics[width = 0.162\columnwidth,height=1.2cm]{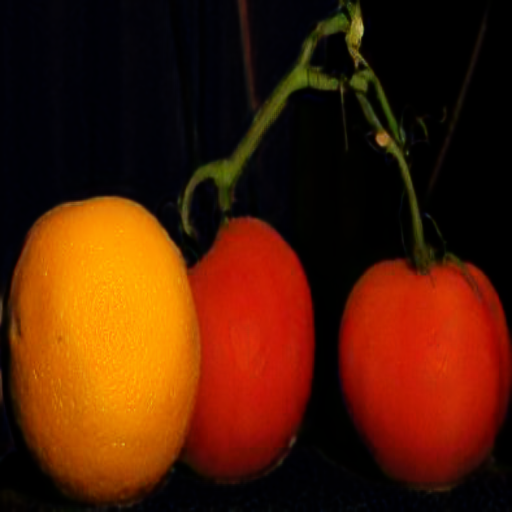}}\hfill\\
	\subfloat{\includegraphics[width = 0.162\columnwidth,height=1.2cm]{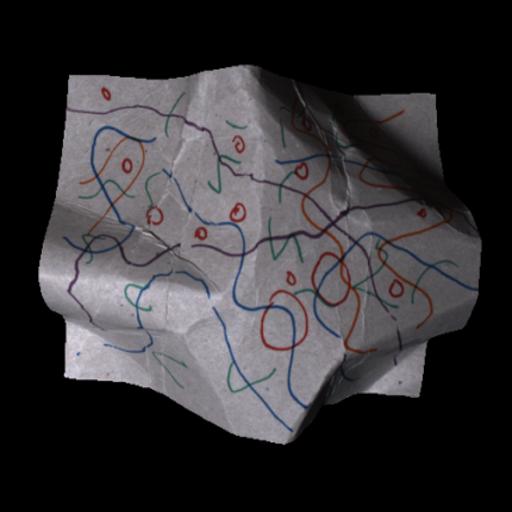}}\hfill
	\subfloat{\includegraphics[width = 0.162\columnwidth,height=1.2cm]{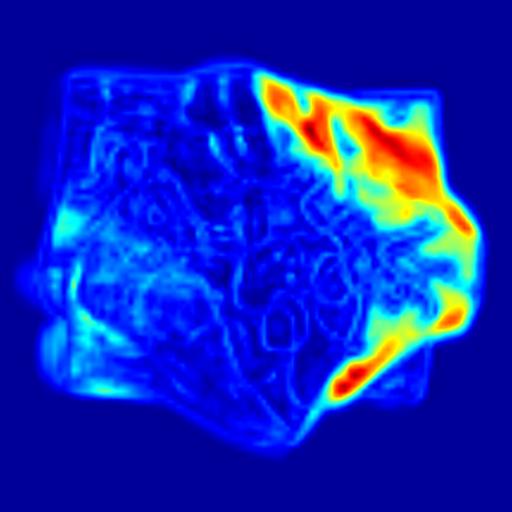}}\hfill
	\subfloat{\includegraphics[width = 0.162\columnwidth,height=1.2cm]{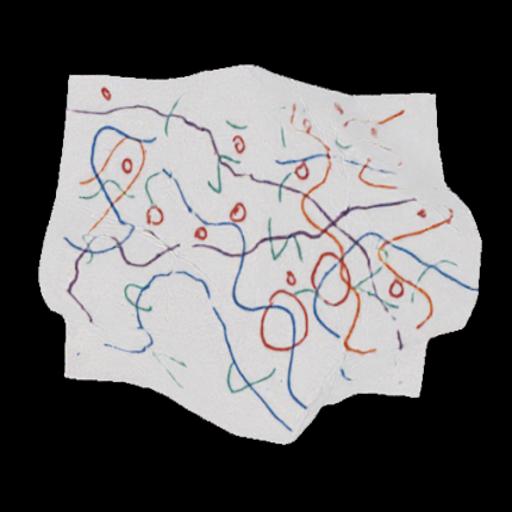}}\hfill	
	\subfloat{\includegraphics[width = 0.162\columnwidth,height=1.2cm]{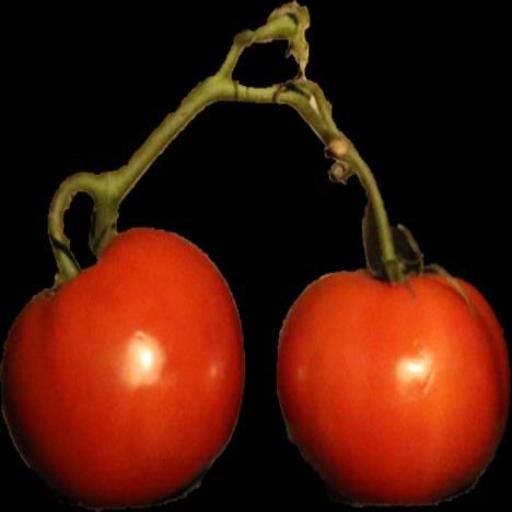}}\hfill
	\subfloat{\includegraphics[width = 0.162\columnwidth,height=1.2cm]{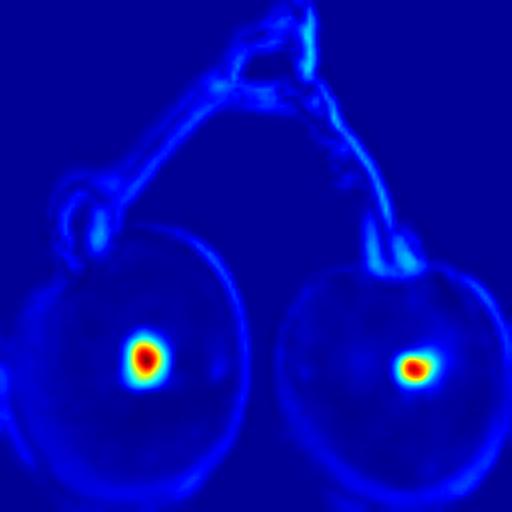}}\hfill
	\subfloat{\includegraphics[width = 0.162\columnwidth,height=1.2cm]{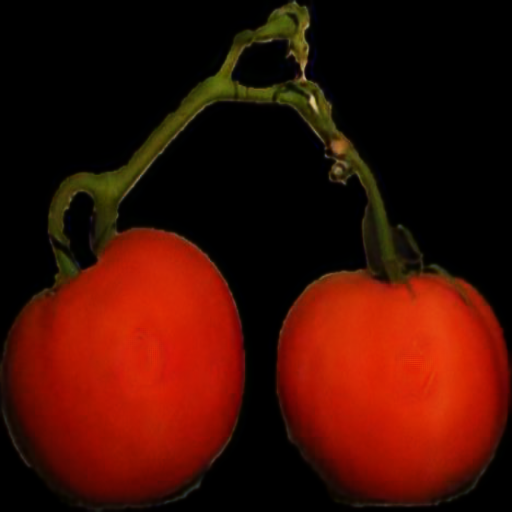}}\hfill\\
	\subfloat{\includegraphics[width = 0.162\columnwidth,height=1.2cm]{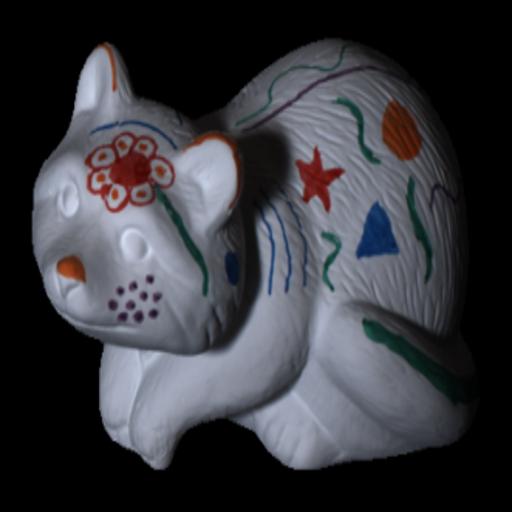}}\hfill
	\subfloat{\includegraphics[width = 0.162\columnwidth,height=1.2cm]{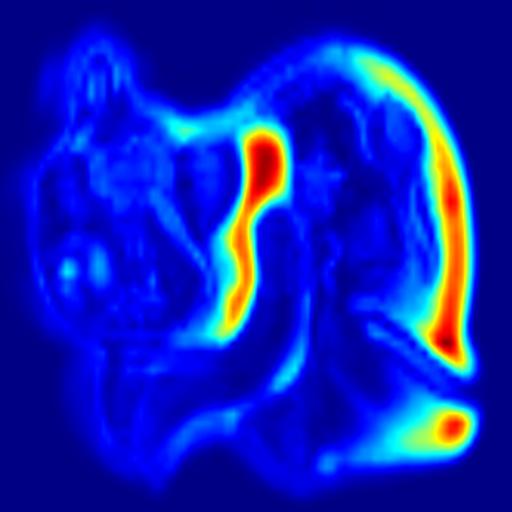}}\hfill
	\subfloat{\includegraphics[width = 0.162\columnwidth,height=1.2cm]{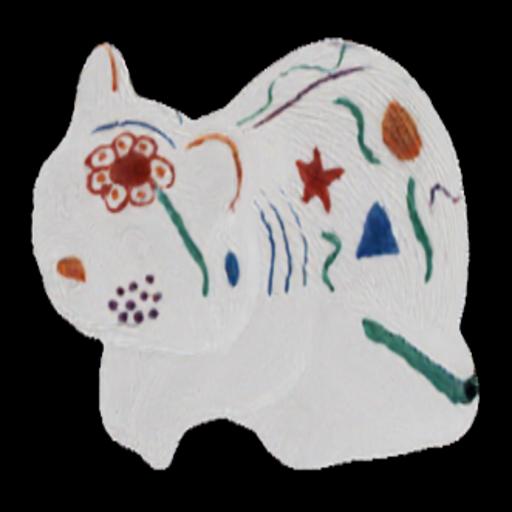}}\hfill
	\subfloat{\includegraphics[width = 0.162\columnwidth,height=1.2cm]{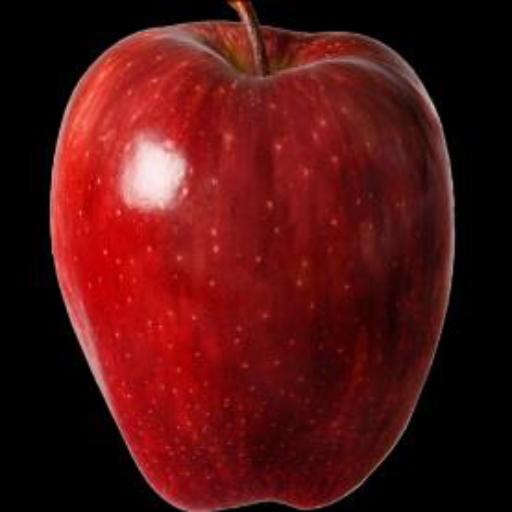}}\hfill
	\subfloat{\includegraphics[width = 0.162\columnwidth,height=1.2cm]{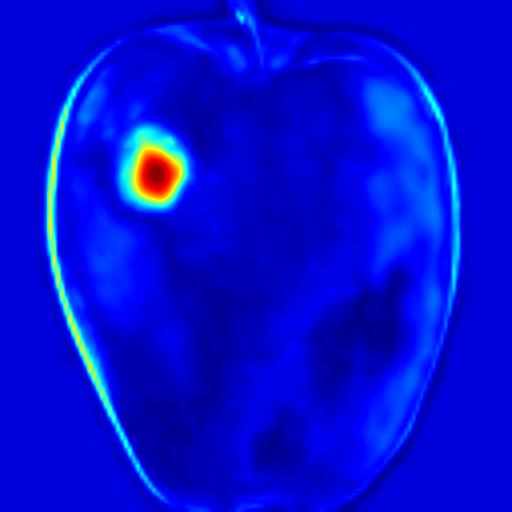}}\hfill
	\subfloat{\includegraphics[width = 0.162\columnwidth,height=1.2cm]{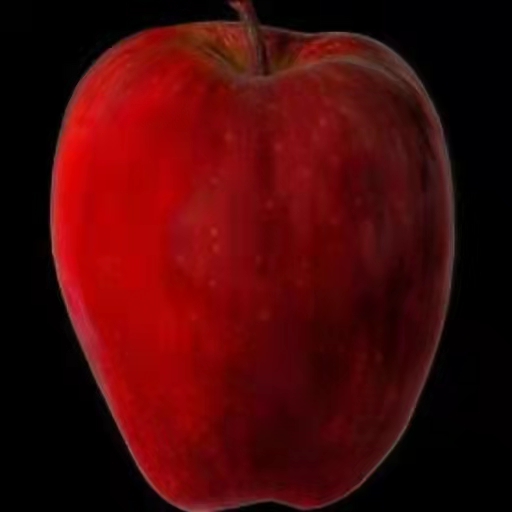}}\hfill\\
	\setcounter{subfigure}{0}
	\subfloat[Input]{\includegraphics[width = 0.162\columnwidth,height=1.2cm]{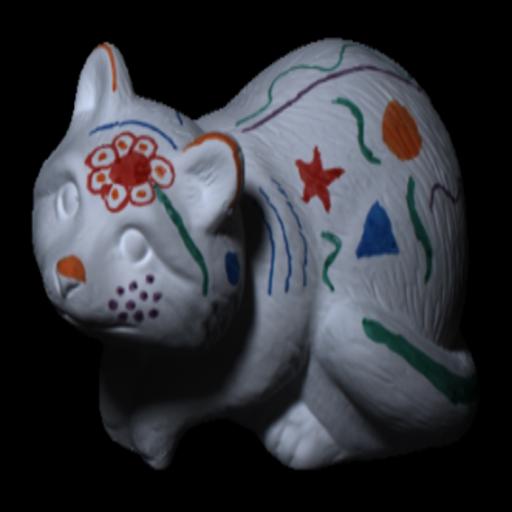}}\hfill
	\subfloat[S-Aware]{\includegraphics[width = 0.162\columnwidth,height=1.2cm]{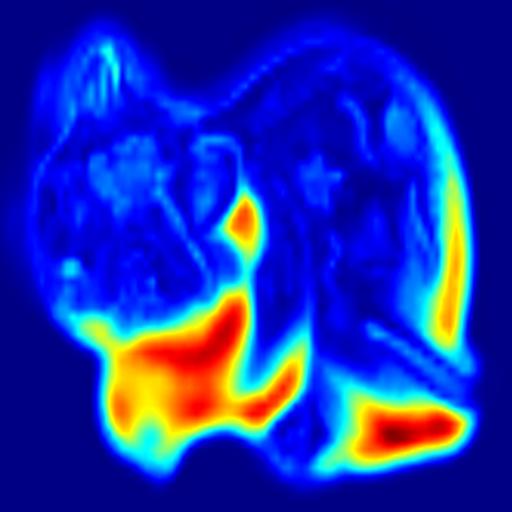}\label{fig:shading_aware}}\hfill
	\subfloat[Output]{\includegraphics[width = 0.162\columnwidth,height=1.2cm]{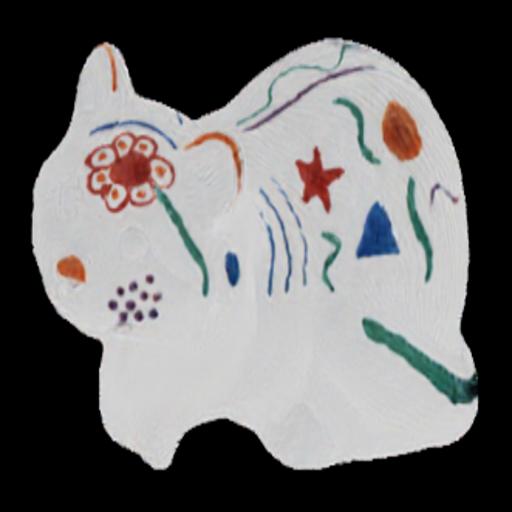}}\hfill			
	\subfloat[Input]{\includegraphics[width = 0.162\columnwidth,height=1.2cm]{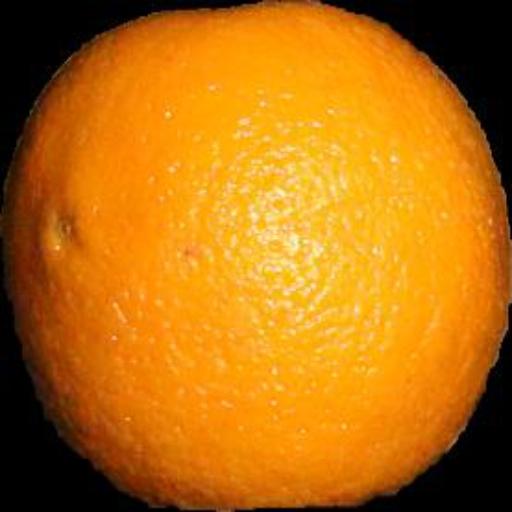}}\hfill
	\subfloat[S-Aware]{\includegraphics[width = 0.162\columnwidth,height=1.2cm]{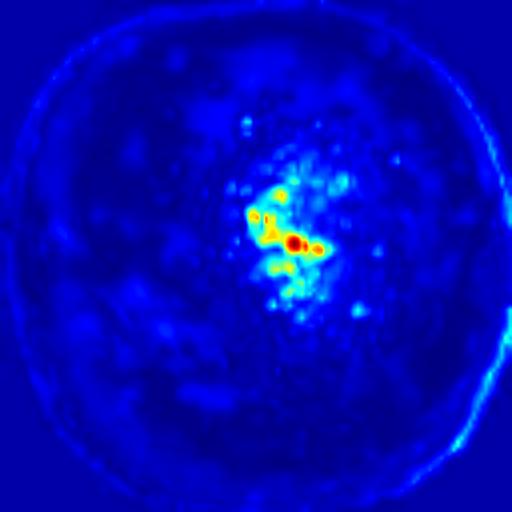}\label{fig:specular_aware}}\hfill
	\subfloat[Output]{\includegraphics[width = 0.162\columnwidth,height=1.2cm]{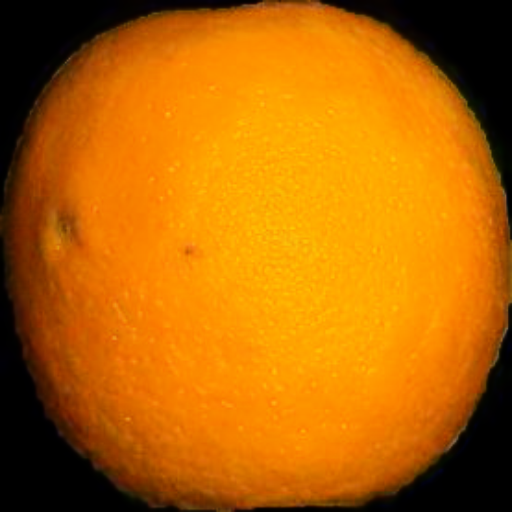}}\hfill\\
	\caption{Left-most three columns are the real images from the MIT dataset~\cite{grosse2009ground}. The right-most three columns are the images from the real highlight~\cite{yi2020leveraging} and synthetic ShapeNet dataset~\cite{chang2015shapenet}. Our S-Aware network focuses on shadow/specular regions, which are the modulated activation features.}
	\label{fig:AwareNet}
\end{figure}

\vspace{0.3cm}
\noindent \textbf{Shading Smooth Loss} 
To impose smooth constraints on the shading layer, we employ a smooth loss that computes the horizontal and vertical gradients:
\begin{equation}
	\mathcal{L}_\mathit{S}^\text{smooth} = |\nabla ({\mathit{S}_i})|_1.
	\label{eq:smoothloss} 
\end{equation}
 
\vspace{0.3cm}
\noindent \textbf{Reflectance Sparse Gradient Loss} 
Unlike the shading layer, the gradients of the reflectance layer are sparse~\cite{land1971lightness}. 
To pursue this, we develop a regularization term made by the reweighted $l_p$ norm~\cite{Reweighted} to punish the gradients:
\begin{align}
	\mathcal{L}_\mathit{R}^\text{sparse} &= \omega |\nabla ({\mathit{R}_i})|_1 ;\quad
	\omega = \frac{1}{|\nabla ({\mathit{R}_i})|^{1-p}+\epsilon},
	\label{eq:sparseloss} 
\end{align}
where $\epsilon$ is a small positive number that stabilizes the numerical
calculation, $p$ is a parameter that determines the extent of the sparsity, $p=0.5$,
and $\omega$ is regarded as a constant that is not involved
in the back-propagation optimization.

We multiply each of the above-mentioned loss functions with its respective weight, where in our experiments, we empirically set $\lambda_\mathit{R}^\text{sf}$, 
$\lambda_\mathit{R}^\text{hf}$, $\lambda_\mathit{R,S}^\text{grad}$ to 1,  since they are in the same scale.
We also empirically set $\lambda_\mathit{S}^\text{smooth}=0.5$, $\mathcal{L}_\mathit{R}^\text{sparse}=0.01$. 

\begin{figure}[t!]
	\centering
	\subfloat{\includegraphics[width=1\linewidth]{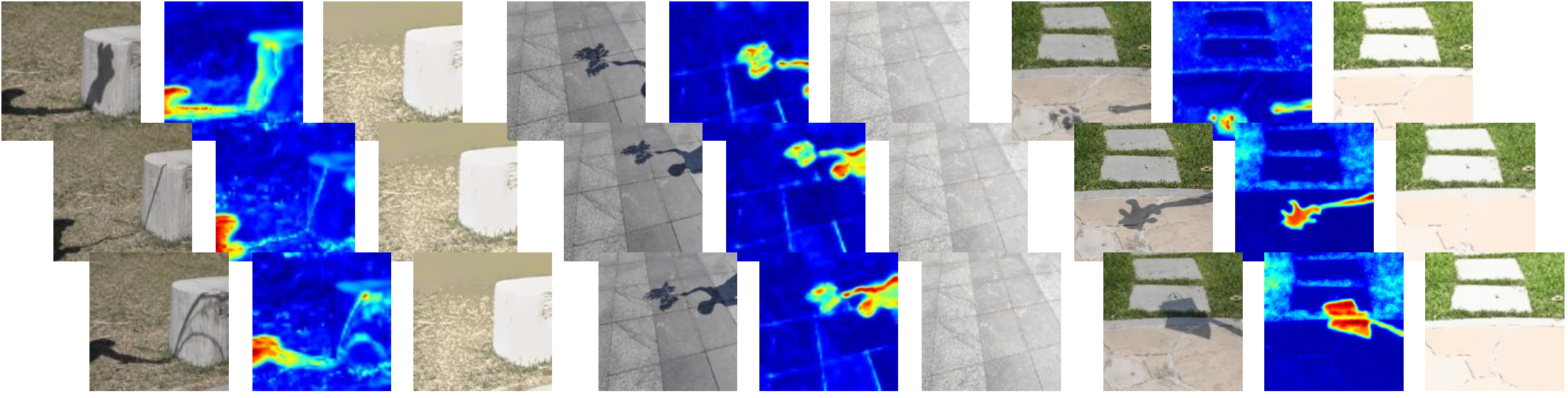}}
	\caption{Our S-Aware network can capture spatially varying regions from dataset~\cite{qu2017deshadownet,wang2018stacked}, the diffuse reflectance of the objects in the static scene always remains the same.}
	\label{fig:shading_variant}	
\end{figure}

\subsection{Shadow/Specular-Aware Network}
To further refine the reflectance layer, in the second stage, our S-Aware network focuses on shadow/specular regions. 
As shown in Fig.~\ref{fig:model}, the architecture of the S-Aware network is represented by $G_s$ (the subscript $s$ stands for shadow/specular).
It has an encoder $E_s$, a decoder $D_s$, and a shadow/specularity classifier (S-Classifier) $C_\text{cls}$.
The input to the S-Aware network is the initial reflectance layer $\mathit{R}_i$ and the original input image $\mathit{I}$.
The output is the refined reflectance image $\mathit{R}_f$, which is our final output.

We train our S-Classifier to predict the two-category classification probability and to judge whether the encoded features come from a certain class.
The classification’s confidence score is the likelihood of the input image belonging to the shadow/shadow-free or specular/specular-free classes.
To be more specific, the encoder $E_s$ generates the encoded features $\mathit{F}$.
Inspired by the class activation mapping~\cite{zhou2016learning,kim2019u,jin2022unsupervised}, we multiply the features with classification weights $\mathit{w}^s$, learned by our S-Classifier. Then, we obtain the shadow/specular-activated features, denoted as $\mathit{A}^s$, defined as:
\begin{equation}
	\mathit{A}^s = \frac{1}{m}\sum_{i=1}^{m}{\mathit{w}^s}_i {\mathit{F}}_i,
	\label{eq:atten}
\end{equation}where $m$ is the total number of feature maps.
The activated features indicate the importance of the regions to distinguish the shadows from shadow-free, or specularities from specular-free.
The S-Classifier $C_\text{cls}$ performs a two-category classification, in which the loss is based on the following equation:
\begin{equation}
	\mathcal{L}_{\text{cls}}=-(\mathbb{E}_{x \sim I}[\log(C_\text{cls}(x))]+\mathbb{E}_{x \sim R_i}[\log(1-C_\text{cls}(x))]).
	\label{eq:loss_cls}
	\nonumber
\end{equation}
For this classification task, the activated features generate a form of attention that focuses on the shadow/specular regions, as shown in Fig.~\ref{fig:AwareNet} and Fig.~\ref{fig:shading_variant}.

\begin{figure*}[t]
	\centering
	\captionsetup[subfigure]{font=small, labelformat=empty}
	\captionsetup[subfloat]{farskip=1pt}
	\subfloat{\includegraphics[width = 0.121\textwidth]{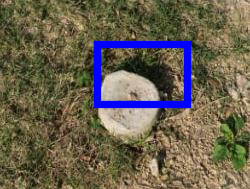}}\hfill
	\subfloat{\includegraphics[width = 0.121\textwidth]{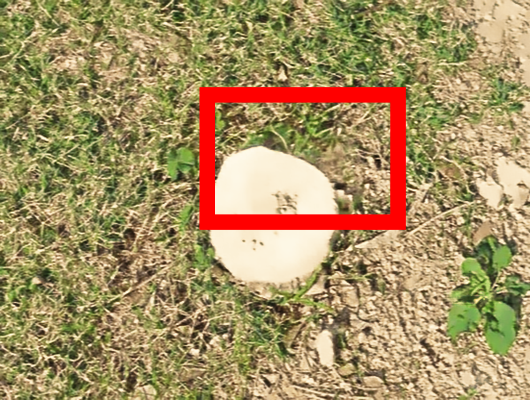}}\hfill
	\subfloat{\includegraphics[width = 0.121\textwidth]{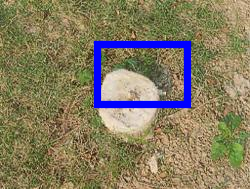}}\hfill
	\subfloat{\includegraphics[width = 0.121\textwidth]{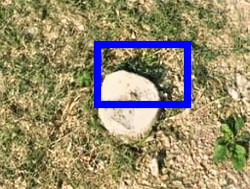}}\hfill
	\subfloat{\includegraphics[width = 0.121\textwidth]{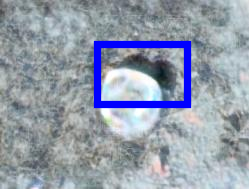}}\hfill
	\subfloat{\includegraphics[width = 0.121\textwidth]{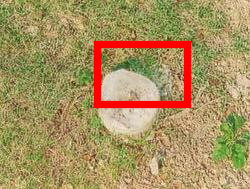}}\hfill
	\subfloat{\includegraphics[width = 0.121\textwidth]{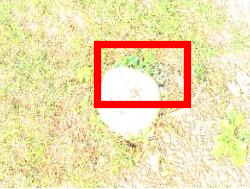}}\hfill
	\subfloat{\includegraphics[width = 0.121\textwidth]{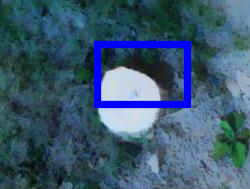}}\hfill\\
	\setcounter{subfigure}{0}
	\subfloat[Input]{\includegraphics[width = 0.121\textwidth]{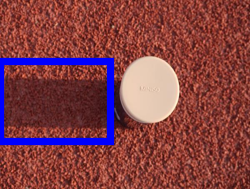}}\hfill
	\subfloat[Ours]{\includegraphics[width = 0.121\textwidth]{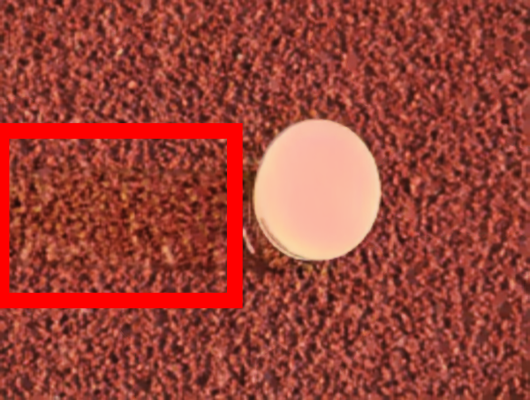}}\hfill
	\subfloat[PIENet~\shortcite{das2022pie}]{\includegraphics[width = 0.121\textwidth]{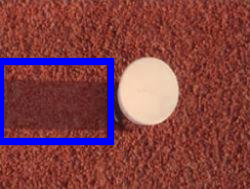}}\hfill
	\subfloat[UIDNet~\shortcite{zhang2021unsupervised}]{\includegraphics[width = 0.121\textwidth]{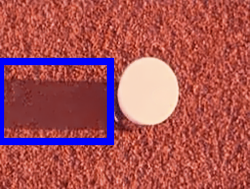}}\hfill
	\subfloat[USI3D~\shortcite{liu2020unsupervised}]{\includegraphics[width = 0.121\textwidth]{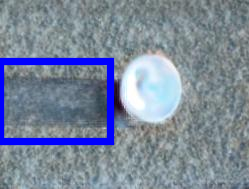}}\hfill
	\subfloat[ShadNet~\shortcite{baslamisli2021shadingnet}]{\includegraphics[width = 0.121\textwidth]{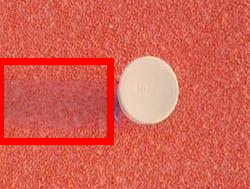}}\hfill
	\subfloat[STAR~\shortcite{xu2020star}]{\includegraphics[width = 0.121\textwidth]{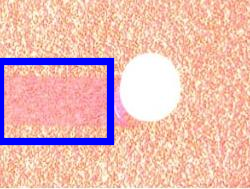}}\hfill
	\subfloat[InvRen~\shortcite{yu2019inverserendernet}]{\includegraphics[width = 0.121\textwidth]{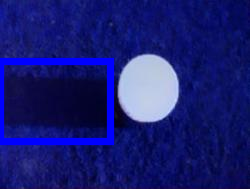}}\hfill
	\caption{Comparing results on the shadow dataset~\cite{hu2019mask}. Most existing methods perform poorly in separating out shadows from the reflectance layer, while our method more effectively estimates the reflectance layer that is free from shadows.}
	\label{fig:shadow_all}
\end{figure*}

\begin{figure*}[t!]
	\centering
	\captionsetup[subfigure]{font=small, labelformat=empty}
	\captionsetup[subfloat]{farskip=1pt}
	\subfloat{\includegraphics[width = 0.121\textwidth]{fig/MIT/box-1_i.jpg}}\hfill
	\subfloat{\includegraphics[width = 0.121\textwidth]{fig/MIT/box-1_gt.jpg}}\hfill
	\subfloat{\includegraphics[width = 0.121\textwidth]{fig/MIT/box-1_our.jpg}}\hfill
	\subfloat{\includegraphics[width = 0.121\textwidth]{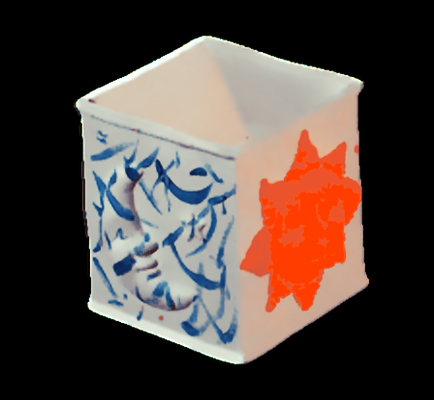}}\hfill
	\subfloat{\includegraphics[width = 0.121\textwidth]{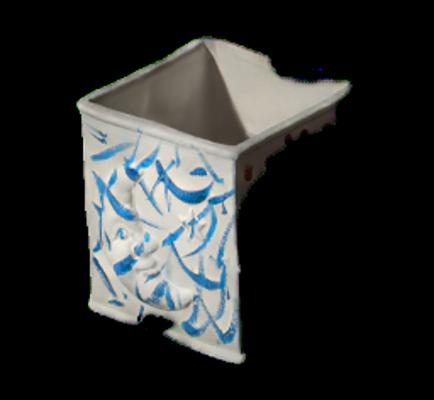}}\hfill
	\subfloat{\includegraphics[width = 0.121\textwidth]{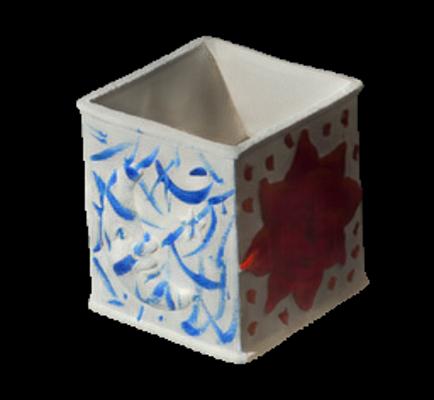}}\hfill
	\subfloat{\includegraphics[width = 0.121\textwidth]{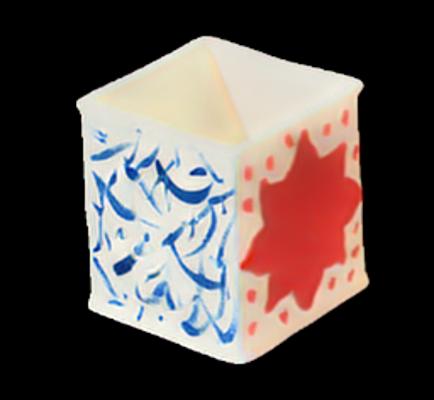}}\hfill
	\subfloat{\includegraphics[width = 0.121\textwidth]{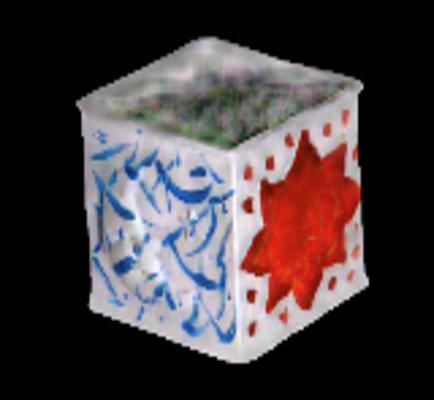}}\hfill\\
	\subfloat[Input]{\includegraphics[width = 0.121\textwidth]{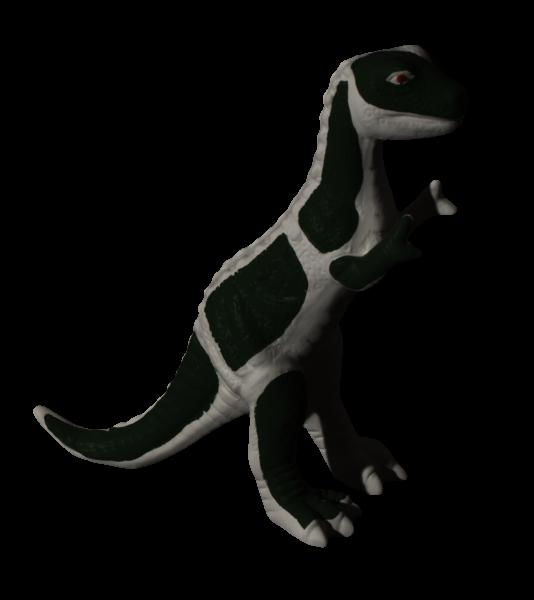}}\hfill
	\subfloat[Ground Truth]{\includegraphics[width = 0.121\textwidth]{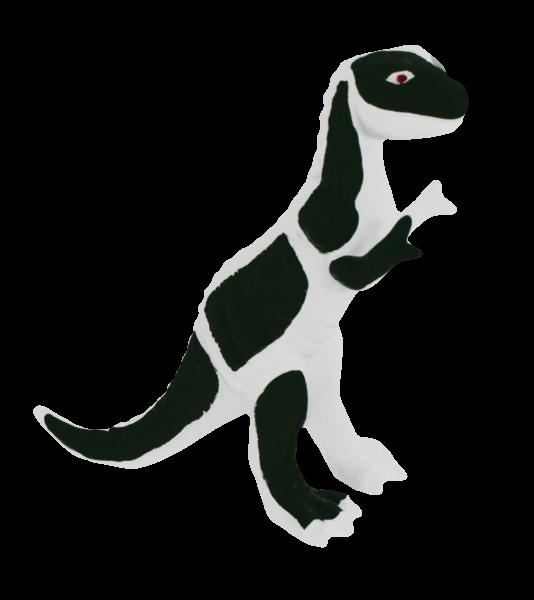}}\hfill
	\subfloat[Ours]{\includegraphics[width = 0.121\textwidth]{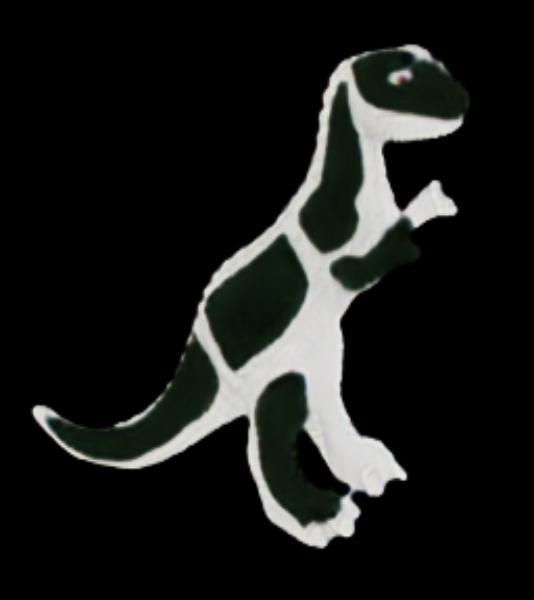}}\hfill
	\subfloat[UIDNet~\shortcite{zhang2021unsupervised}]{\includegraphics[width = 0.121\textwidth]{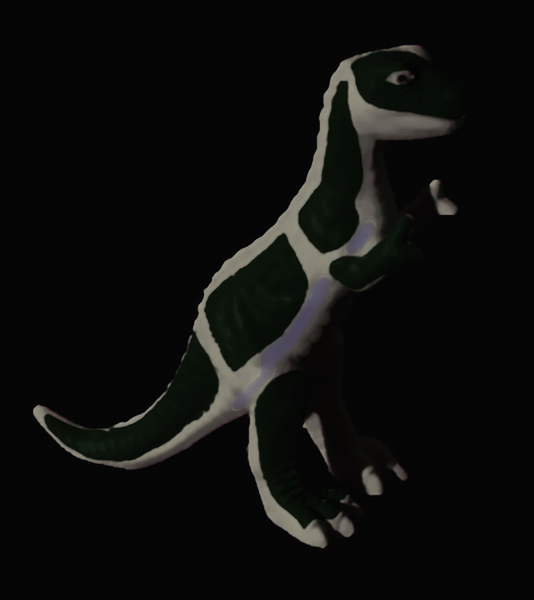}}\hfill
	\subfloat[PIENet~\shortcite{das2022pie}]{\includegraphics[width = 0.121\textwidth]{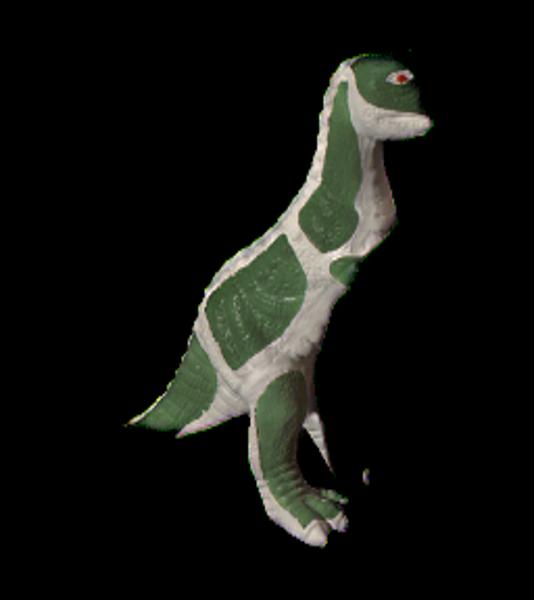}}\hfill
	\subfloat[ShadNet~\shortcite{baslamisli2021shadingnet}]{\includegraphics[width = 0.121\textwidth]{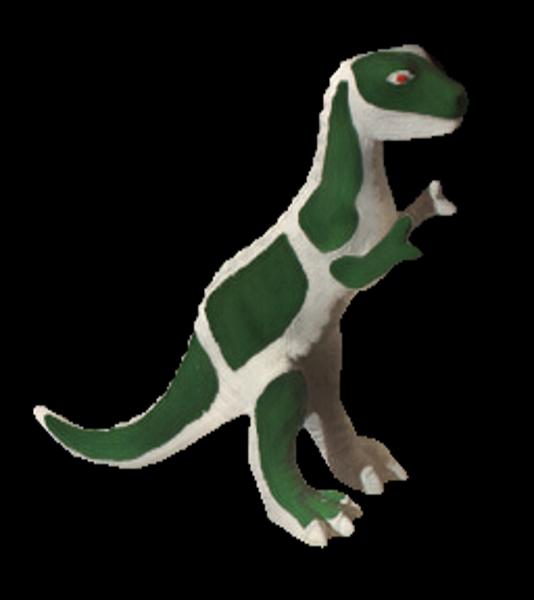}}\hfill
	\subfloat[Physics~\shortcite{baslamisli2021physics}]{\includegraphics[width = 0.121\textwidth]{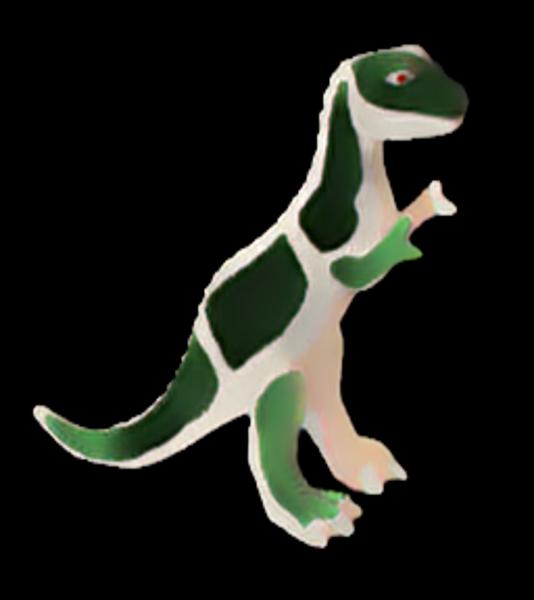}}\hfill
	\subfloat[IntNet~\shortcite{baslamisli2018cnn}]{\includegraphics[width = 0.121\textwidth]{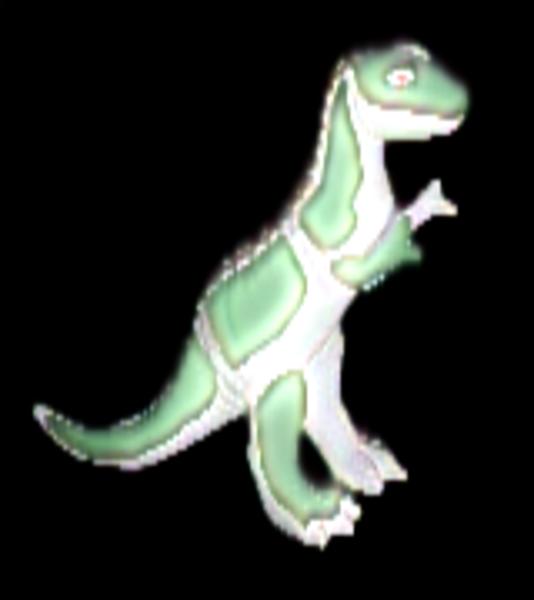}}\hfill
	\caption{Comparing results on the MIT dataset~\cite{grosse2009ground}. Most existing methods perform poorly in separating out shadows from the reflectance layer, while our method more effectively estimates the reflectance layer that is free from shadows.}
	\label{fig:mit}
\end{figure*}

\vspace{0.3cm}
\noindent \textbf{Spatially Varying Regions}
In a spatially varying dataset shown in Fig.~\ref{fig:shading_variant}, the shadow/specular regions of the scenes constantly change, while the diffuse reflectance of the objects in the static scene always remains the same.
To help our network focus on the large shadow/specular regions in the spatially varying dataset to remove them, we use feature statistics and the normalization layer to capture the spatial-variant property of shadows/specularities.
Since the mean and variance of the features encode the statistics of an image, aligning them implies transferring the style of the image. 

Instance Normalization (IN)~\cite{Huang_2017_ICCV} is used in unsupervised style transfer~\cite{liu2020unsupervised}: 
$\phi^{c}=\frac{\phi-\mu^{c}(\phi)}{\sqrt{\sigma^{c}(\phi)^{2}+\epsilon}}$, 
where $\mu^{c}$ and $\sigma^{c}$ are the channel-wise mean and standard deviation, and $\phi$ is the activation of the previous convolutional layer. 
Unlike IN, layer normalization (LN)~\cite{ba2016layer} captures the average and variance of each pixel among a batch of input data:
$\phi^{l}=\frac{\phi-\mu^{l}(\phi)}{\sqrt{\sigma^{l}(\phi)^{2}+\epsilon}}$,
where $\mu^{l}$ and $\sigma^{l}$ are the layer-wise mean and the standard deviation.
Our S-Aware network dynamically adjusts the ratio  $\nu$ $(\mbox{where } \nu \in\{0,1\})$ between IN and LN operations:
$\operatorname{LIN}(\gamma, \beta, \nu)=\gamma\left((1-\nu) \cdot \phi^{c}+\nu \cdot \phi^{l}\right)+\beta$,
where $\gamma$ and $\beta$ are parameters generated by the fully connected layer. $\phi^{c}$, $\phi^{l}$ are the channel-wise, layer-wise normalization functions, respectively.

\vspace{0.3cm}
\noindent \textbf{Adversarial and Translation Losses} Our S-Aware network, $G_s$, is coupled with a discriminator $D_{sf}$. 
We use the least-square GAN (LSGAN) adversarial losses to stabilize our network training:
\begin{equation}
	\mathcal{L}_{\rm adv}(G_s, D_{sf})=  
	\mathbb{E}[\big(D_{sf}(\mathit{I})-1\big)^2]\label{eq:loss_gan1}
	+\mathbb{E}[\big(D_{sf}(\mathit{R}_f))\big)^2].
	\nonumber
\end{equation}
When the S-Aware network, $G_s$, takes the original input, it outputs the reflectance layer.
We enforce the following loss:
\begin{equation}
	\mathcal{L}_{\rm trans}(G_s) = |G_s(\mathit{I})-\mathit{R}_f|_{1}.
	\label{eq:loss_trans}
\end{equation} 

\vspace{0.3cm}
\noindent \textbf{Diffuse Loss} 
We learn the  reflectance layer also from the diffuse reflection component $\mathit{I}_d$, where  $\mathit{I}_d=\mathit{R}_d \odot \mathit{S}$ with $\mathit{R}_d$ is the true diffuse reflectance layer.
For images suffered only from shadows, the diffuse reflection component is the image itself, where $\mathit{I}_d=\mathit{I}$.
For images suffered from specular highlights, the diffuse reflection component can be computed: $\mathit{I}_d=\mathit{I}-\mathit{I}_s$~\cite{shafer1985using,tan2005specular}, where $\mathit{I}_s$ is the specular reflection component. 
Some real~\cite{yi2020leveraging} and synthetic datasets~\cite{shi2017learning} provide the information of the diffuse reflection component and the corresponding image together, which can be used in our training.
Hence, we define our diffuse loss as:
\begin{equation}
\mathcal{L}_{\rm diff}(G_s) = |\mathit{R}_f \odot \mathit{S}_i-\mathit{I}_d|_1.
\label{eq:reconloss}
\end{equation}

\vspace{0.2cm}
\noindent \textbf{Overall Loss}
We multiply each loss function with its respective weight, and sum them together to obtain our overall loss function. 
The weights of the losses, $\mathcal{L}_{\rm cls}$, $\mathcal{L}_{\rm adv}$, $\mathcal{L}_{\rm trans}$ are empirically set to 5, 1, 5 in our experiments. The weights of $\mathcal{L}_{\rm diff}$ are 1, since they are in the same scale.

\begin{figure*}[t!]
	\captionsetup[subfigure]{font=small, labelformat=empty}
	\captionsetup[subfloat]{farskip=1pt}
	\centering
	\subfloat{\includegraphics[width = 0.097\textwidth]{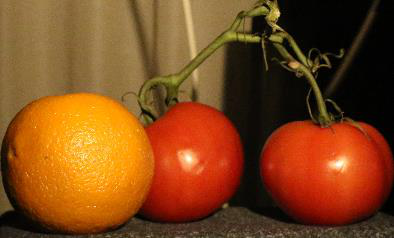}}\hfill
	\subfloat{\includegraphics[width = 0.097\textwidth]{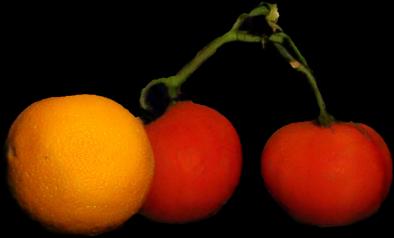}}\hfill
	\subfloat{\includegraphics[width = 0.097\textwidth]{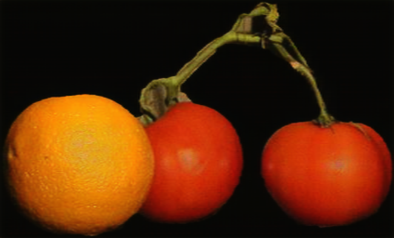}}\hfill
	\subfloat{\includegraphics[width = 0.097\textwidth]{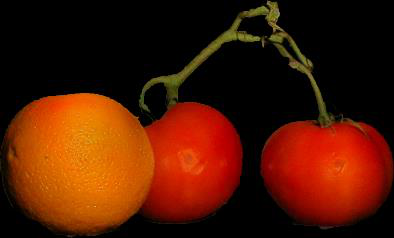}}\hfill
	\subfloat{\includegraphics[width = 0.097\textwidth]{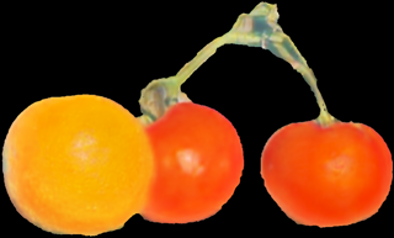}}\hfill
	\subfloat{\includegraphics[width = 0.097\textwidth]{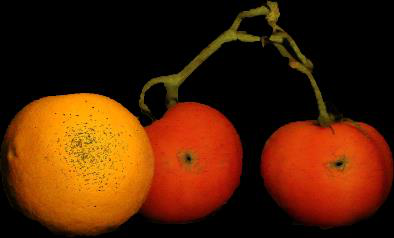}}\hfill
	\subfloat{\includegraphics[width = 0.097\textwidth]{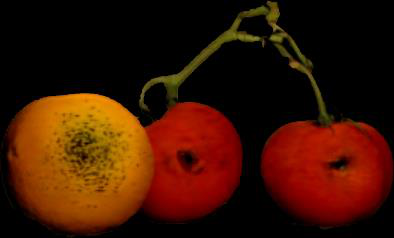}}\hfill
	\subfloat{\includegraphics[width = 0.097\textwidth]{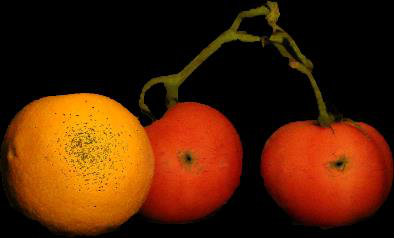}}\hfill
	\subfloat{\includegraphics[width = 0.097\textwidth]{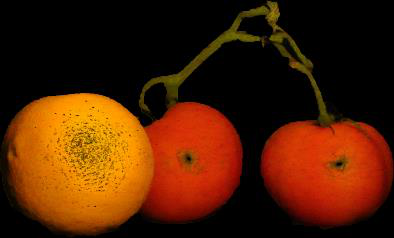}}\hfill
	\subfloat{\includegraphics[width = 0.097\textwidth]{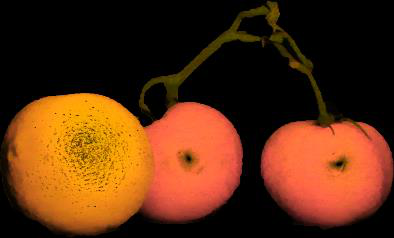}}\hfill\\
	\subfloat[Input]{\includegraphics[width = 0.097\textwidth,height=1.8cm]{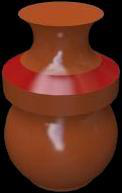}}\hfill
	\subfloat[Ours]{\includegraphics[width = 0.097\textwidth,height=1.8cm]{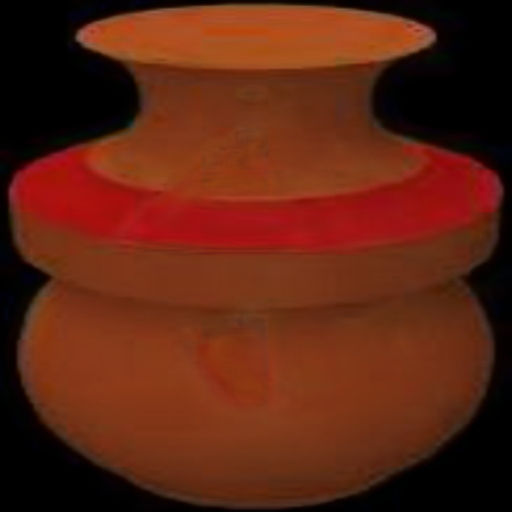}}\hfill
	\subfloat[Fu~\shortcite{fu2021multi}]{\includegraphics[width = 0.097\textwidth,height=1.8cm]{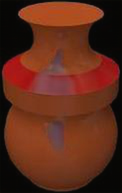}}\hfill
	\subfloat[Yi~\shortcite{yi2020leveraging}]{\includegraphics[width = 0.097\textwidth,height=1.8cm]{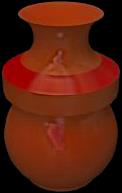}}\hfill
	\subfloat[Li~\shortcite{li2020inverse}]{\includegraphics[width = 0.097\textwidth,height=1.8cm]{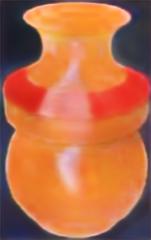}}\hfill
	\subfloat[Guo~\shortcite{guo2018single}]{\includegraphics[width = 0.097\textwidth,height=1.8cm]{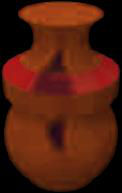}}\hfill
	\subfloat[ShapeNet]{\includegraphics[width = 0.097\textwidth,height=1.8cm]{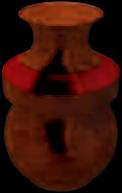}}\hfill
	\subfloat[Shen~\shortcite{shen2013real}]{\includegraphics[width = 0.097\textwidth,height=1.8cm]{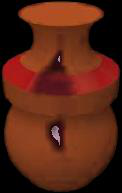}}\hfill
	\subfloat[Yang~\shortcite{yang2010real}]{\includegraphics[width = 0.097\textwidth,height=1.8cm]{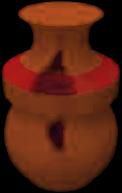}}\hfill
	\subfloat[Tan~\shortcite{tan2005specular}]{\includegraphics[width = 0.097\textwidth,height=1.8cm]{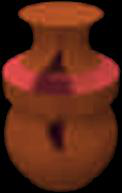}}\hfill
	\caption{Comparing results on the real highlight dataset~\cite{yi2020leveraging} and synthetic specular ShapeNet dataset~\cite{shi2017learning}. Existing methods perform poorly in separating specularities from the reflectance layer.}
	\label{fig:hl}
\end{figure*}

\begin{table}[t]
	\centering
	\resizebox{0.48\textwidth}{!}{
			\begin{tabular}{llc}\hline
			Methods & Training set & WHDR\%$\downarrow$ \\\hline
			STAR~\shortcite{xu2020star}       & $-$ & $32.90$ \\
			IIW~\shortcite{bell2014intrinsic} & $-$ & $20.64$ \\\hline
			ShapeNet~\shortcite{shi2017learning}    & ShapeNet (2.5M) & $59.40$ \\
			IntrinsicNet~\shortcite{baslamisli2018cnn}       & ShapeNet (20K) & $32.10$ \\
			CGI~\shortcite{li2018cgintrinsics}               & SUNCG & $26.10$ \\
			InvRen~\shortcite{yu2019inverserendernet} & MegaDepth & $21.40$ \\
			PIE-Net~\shortcite{das2022pie}                & NED+IIW & $18.50$ \\
			Fan~\textit{et al.}~\shortcite{fan2018revisiting}*\footnote{}& IIW &\bf{14.45} \\\hline
			Yi~\shortcite{yi2020leveraging} & CustomerPhotos & $51.10$ \\
			Physics~\shortcite{baslamisli2021physics}  & ShapeNet (20K) & $28.90$ \\
			Ma~\textit{et al.}~\shortcite{ma2018single}   & IIW	    & $28.04$ \\
			IIDWW~\shortcite{li2018learning}              & BigTime & $20.30$ \\
			USI3D~\shortcite{liu2020unsupervised}      & CGI+IIW & $18.69$ \\
			UIDNet~\shortcite{zhang2021unsupervised}        & $-$ & $18.21$ \\\hline
			Ours                                    & IIW  &\bf{17.97}  \\\hline\end{tabular}
		}
	\caption{Results for IIW dataset~\cite{bell2014intrinsic} using WHDR~\cite{bell2014intrinsic}. * denotes use ground truth supervision in training and post-processed to benefit the WHDR score.}
	\label{tb:iiw}
\end{table}

\section{Experimental Results}
\label{sec:experiments}
\begin{figure}[t!]
	\captionsetup[subfigure]{font=small, labelformat=empty}
	\captionsetup[subfloat]{farskip=1pt}
	\centering
	\subfloat{\includegraphics[width = 0.195\columnwidth]{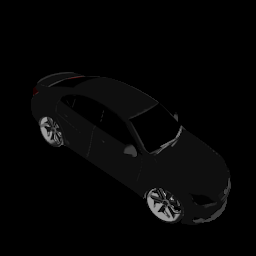}}\hfill
	\subfloat{\includegraphics[width = 0.195\columnwidth]{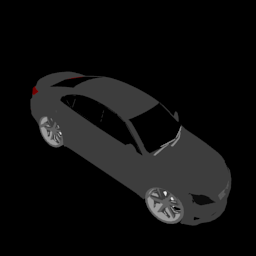}}\hfill
	\subfloat{\includegraphics[width = 0.195\columnwidth]{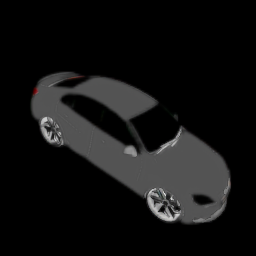}}\hfill
	\subfloat{\includegraphics[width = 0.195\columnwidth]{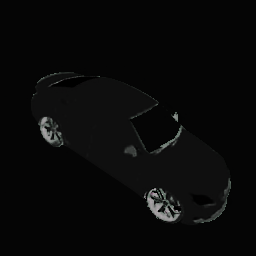}}\hfill
	\subfloat{\includegraphics[width = 0.195\columnwidth]{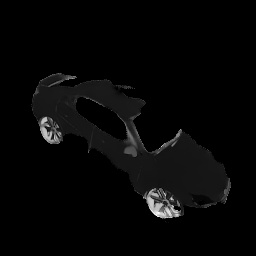}}\hfill	
	\subfloat[Input]{\includegraphics[width = 0.195\columnwidth]{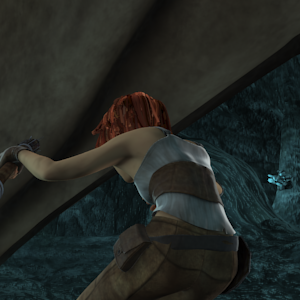}}\hfill
	\subfloat[GT]{\includegraphics[width = 0.195\columnwidth]{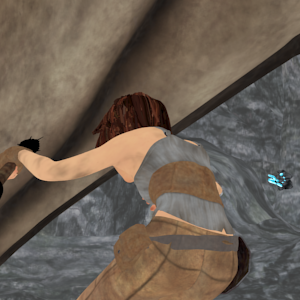}}\hfill
	\subfloat[Ours]{\includegraphics[width = 0.195\columnwidth]{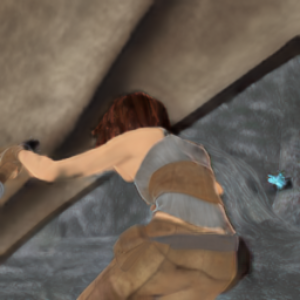}}\hfill
	\subfloat[UIDNet]{\includegraphics[width = 0.195\columnwidth]{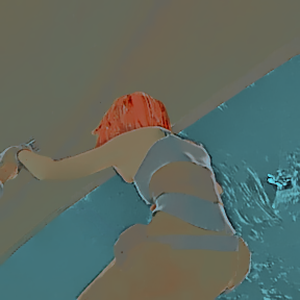}}\hfill
	\subfloat[PIENet]{\includegraphics[width = 0.195\columnwidth]{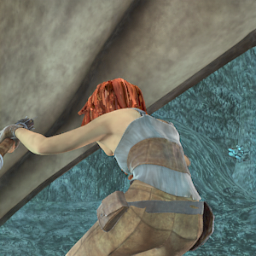}}\hfill
	\caption{Comparing results on the MPI-Sintel~\cite{butler2012naturalistic} (last row) and ShapeNet intrinsic dataset~\cite{shi2017learning} (first row).} 
	\label{fig:syn}
\end{figure}

\noindent \textbf{Reflectance Layer Estimation on Shadows}
To evaluate our method, we use 2 real (MIT Intrinsic, IIW) and 2 synthetic intrinsic image decomposition datasets (MPI-Sintel, ShapeNet), and 2 shadow datasets (SRD, USR).
Fig.~\ref{fig:shadow_all} and Fig.~\ref{fig:mit} show our results on the real image USR shadow~\cite{hu2019mask} and MIT Intrinsic dataset~\cite{grosse2009ground}. 
Fig.~\ref{fig:syn} shows our results on the synthetic datasets MPI-Sintel~\cite{butler2012naturalistic} and ShapeNet~\cite{chang2015shapenet}.
Most SOTA methods perform poorly in separating shadows from the reflectance layer. 
When the images have shadow regions, these methods cannot estimate the information under the shadow regions.
Our results are free from shadows.

We make a fair comparison with SOTA intrinsic image decomposition methods:
PIENet~\cite{das2022pie}, UIDNet~\cite{zhang2021unsupervised}, ShadingNet~\cite{baslamisli2021shadingnet}, Physics~\cite{baslamisli2021physics}, USI3D~\cite{liu2020unsupervised};
optimization-based method: STAR~\cite{xu2020star}, 
inverse rendering method: InverseRenNet~\cite{yu2019inverserendernet}, etc.

\vspace{0.2cm}
\noindent \textbf{Reflectance Layer Estimation on Specularities}
To evaluate our method, we use 1 synthetic intrinsic image decomposition dataset and 1 real specular dataset.
Fig.~\ref{fig:hl} shows our results on real~\cite{yi2020leveraging} and synthetic specular ShapeNet~\cite{shi2017learning} datasets.
Our method estimates the reflectance layer more effectively, in particular for the specular regions, while others may fail to recover some parts of the specular regions, leaving black artefacts in the specular/saturated regions.

To fairly compare to specular methods, we include learning-based intrinsic image decomposition methods:
Yi~\cite{yi2020leveraging}, ShapeNet~\cite{shi2017learning},
inverse rendering method: Li~\cite{li2020inverse}, etc.
We also include learning-based highlight separation method~\cite{fu2021multi},
optimization-based methods~\cite{guo2018single,shen2013real,yang2010real,tan2005specular}.

\vspace{0.2cm}
\noindent \textbf{Quantitative Evaluation}
Tables~\ref{tb:iiw},~\ref{tb:shapenet} show quantitative results on IIW and ShapeNet datasets.
In Table~\ref{tb:iiw}, for the quantitative evaluation of reflectance images, we employ the weighted human disagreement rate (WHDR) from~\cite{bell2014intrinsic}.
In Table~\ref{tb:shapenet}, we employ si-MSE, si-LMSE from~\cite{barron2014shape}. 

\begin{figure}[t!]
	\centering
	\captionsetup[subfigure]{font=small, labelformat=empty}
	\captionsetup[subfloat]{farskip=1pt}
	\setcounter{subfigure}{0}
	\subfloat[Input]{\includegraphics[width = 0.162\columnwidth]{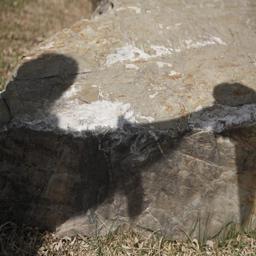}}\hfill
	\subfloat[Stage1]{\includegraphics[width = 0.162\columnwidth]{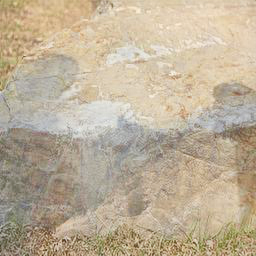}}\hfill
	\subfloat[Output]{\includegraphics[width = 0.162\columnwidth]{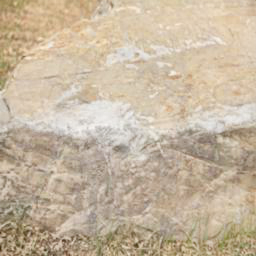}}\hfill
	\subfloat[Input]{\includegraphics[width = 0.162\columnwidth]{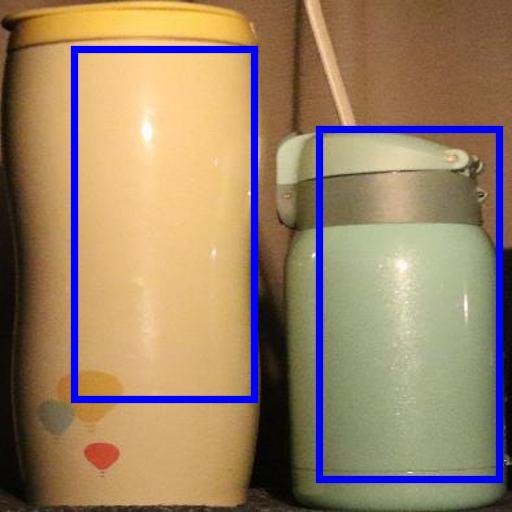}}\hfill
	\subfloat[Stage1]{\includegraphics[width = 0.162\columnwidth]{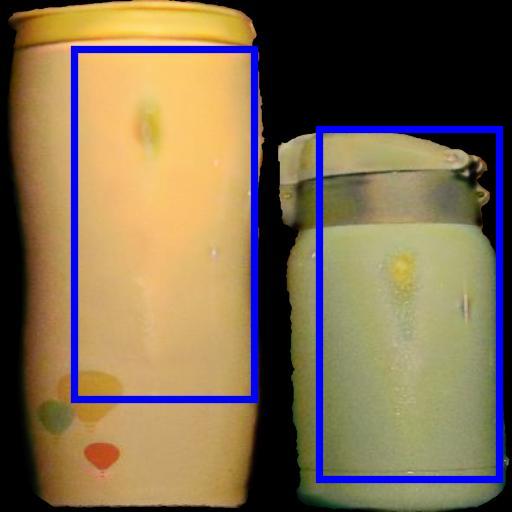}}\hfill
	\subfloat[Output]{\includegraphics[width = 0.162\columnwidth]{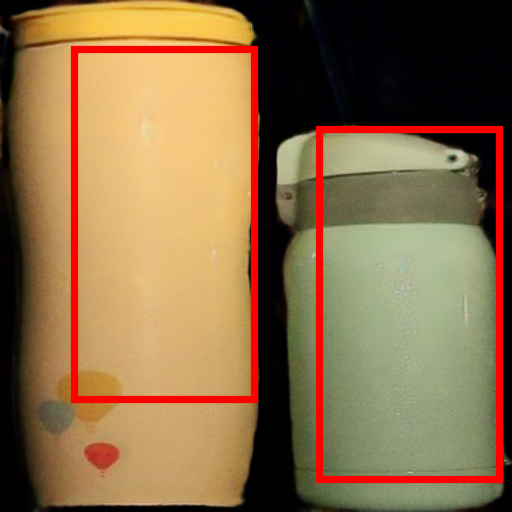}}\hfill\\
	\caption{Ablation studies on stage-one.}
	\label{fig:ablation}
\end{figure}

\vspace{0.2cm}
\noindent \textbf{Ablation Studies}
We conduct ablation studies to analyze the effectiveness of our first-stage reflectance layer guidance, which is shown in Fig.~\ref{fig:ablation}.
To show the effectiveness of the first stage, we train our network without the first stage.
We directly input the image to the second stage, bypassing the first stage.
Table~\ref{tb:shapenet} shows that our first stage has a performance gain of 35\% (from 0.85 to 0.63) on the ShapeNet dataset; implying our first stage provides a good initial shadow/specular-free guidance.
For the first stage, we further study the effectiveness of shadow-free loss ($\mathcal{L}_\mathit{R}^\text{sf}$) and specular-free loss ($\mathcal{L}_\mathit{R}^\text{hf}$).
We use shadow/specular-free images in our losses to constrain our initial estimate. We also use other losses to constrain our initial estimate. Hence, the output of our first stage can outperform specular/shadow-free images.
For the second stage, we compare our results with and without S-Aware design.
That means we remove the S-Classifier and the classification loss.
The corresponding quantitative results are shown in Table~\ref{tb:shapenet}.

\begin{table}[t]
	\centering
	\resizebox{0.48\textwidth}{!}{
		\begin{tabular}{l|ccc|c}\hline
		& \multicolumn{3}{c|}{si-MSE$\downarrow$}       &si-LMSE$\downarrow$   \\ \hline
		Methods                                                                       & R			  & S	    & Avg.    & Total  \\ \hline
		LM~\shortcite{li2014single}                    & 3.38      & 2.96  & 3.17  & 6.23 \\\hline
		Fan~\textit{et al.}~\shortcite{fan2018revisiting} & 3.02      & 3.15  & 3.09  & 7.17 \\\hline
		Ma~\textit{et al.}~\shortcite{ma2018single}       & 2.84      & 2.62  & 2.73  & 5.44 \\
		USI3D~\shortcite{liu2020unsupervised}          & 1.85    &\bf{1.08}  & 1.47  &4.65 \\\hline 
		Ours w/o stage2							    & 1.13      &2.00     & 1.57  &4.93\\
		Ours w/o stage1					            & 0.85      &2.17     & 1.51  &4.85\\
		Ours w/o S-Aware	                            & 0.79      &2.05     & 1.42  &4.15\\ 
		Ours w/o $\mathcal{L}_\mathit{R}^\text{sf}$	& 0.73      &2.07     & 1.40  &4.73\\   
		Ours w/o $\mathcal{L}_\mathit{R}^\text{hf}$	& 0.66      &2.04     & 1.35  &4.19 \\ \hline
		Ours                                &\bf{0.63}    &2.00   &\bf{1.31}  &\bf{4.14}  \\ \hline
		\end{tabular}}
	\caption{Results for ShapeNet intrinsic dataset~\cite{shi2017learning} and the ablation studies of our method.}
	\label{tb:shapenet}
\end{table}

\section{Conclusion}
\label{sec:conclusion}
In this paper, we have proposed a two-stage network to estimate the reflectance layer free from shadows and specularities.
With the reflectance guidance and the S-Aware network, our method can robustly separate shadows and specularities out of the reflectance layer.
We propose novel shadow-free and specular-free losses to estimate the initial reflectance layer. 
To further refine the reflectance layer, we integrate a classifier into our network, enabling our method to focus on shadow/specular regions.
Experimental results have confirmed that our method is effective and outperforms the state-of-the-art reflectance layer estimation methods.

\section*{Acknowledgements}
This research/project is supported by the National Research Foundation, Singapore under its AI Singapore Programme (AISG Award No: AISG2-PhD/2022-01-037[T]), and partially supported by MOE2019-T2-1-130. 
Robby T. Tan's work is supported by MOE2019-T2-1-130.
This research work is also partially supported by the Basic and Frontier Research Project of PCL and the Major Key Project of PCL.
\bibliography{egbib}

\end{document}